\documentclass[10pt,twocolumn,letterpaper]{article}
\usepackage{bm}
\usepackage{multirow}
\usepackage{bbding} 
\usepackage{pifont} 
\usepackage{amsthm,amsmath,amssymb}
\usepackage{mathrsfs}

\usepackage{cvpr}              

\usepackage[dvipsnames]{xcolor}

\definecolor{cvprblue}{rgb}{0.21,0.49,0.74}
\usepackage[pagebackref,breaklinks,colorlinks,citecolor=cvprblue]{hyperref}

\def\paperID{6208} 
\def\confName{CVPR}
\def\confYear{2024}

\definecolor{yxy}{rgb}{0.21,0.49,0.74}

\title{Geometry-aware Reconstruction and Fusion-refined Rendering \\ for Generalizable Neural Radiance Fields}

\author{Tianqi Liu \quad
Xinyi Ye \quad
Min Shi \quad
Zihao Huang \quad
Zhiyu Pan \quad
Zhan Peng \quad
Zhiguo Cao\thanks{Corresponding author} \\
School of AIA, Huazhong University of Science and Technology \\
{\tt\small \{tq\_liu,xinyiye,min\_shi,zihaohuang,zhiyupan,peng\_zhan,zgcao\}@hust.edu.cn}
}

\def\ours{GeFu }
\def\oursnospace{GeFu}

\begin{document}
\maketitle

\begin{abstract}

Generalizable NeRF aims to synthesize novel views for unseen scenes.
Common practices involve constructing variance-based cost volumes for geometry reconstruction and encoding 3D descriptors for decoding novel views.
However, existing methods show limited generalization ability in challenging conditions due to inaccurate geometry, sub-optimal descriptors, and decoding strategies.
We address these issues point by point. First, we find the variance-based cost volume exhibits failure patterns as the features of pixels corresponding to the same point can be inconsistent across different views due to occlusions or reflections. We introduce an Adaptive Cost Aggregation (ACA) approach to amplify the contribution of consistent pixel pairs and suppress inconsistent ones.
Unlike previous methods that solely fuse 2D features into descriptors, our approach introduces a Spatial-View Aggregator (SVA) to incorporate 3D context into descriptors through spatial and inter-view interaction.
When decoding the descriptors, we observe the two existing decoding strategies excel in different areas, which are complementary. A Consistency-Aware Fusion (CAF) strategy is proposed to leverage the advantages of both.
We incorporate the above ACA, SVA, and CAF into a coarse-to-fine framework, termed Geometry-aware Reconstruction and Fusion-refined Rendering (GeFu).
GeFu attains state-of-the-art performance across multiple datasets.
Code is available at \href{https://github.com/TQTQliu/GeFu}{https://github.com/TQTQliu/GeFu}.


\end{abstract}

\begin{figure}
    \centering
    \includegraphics[width=0.47\textwidth]{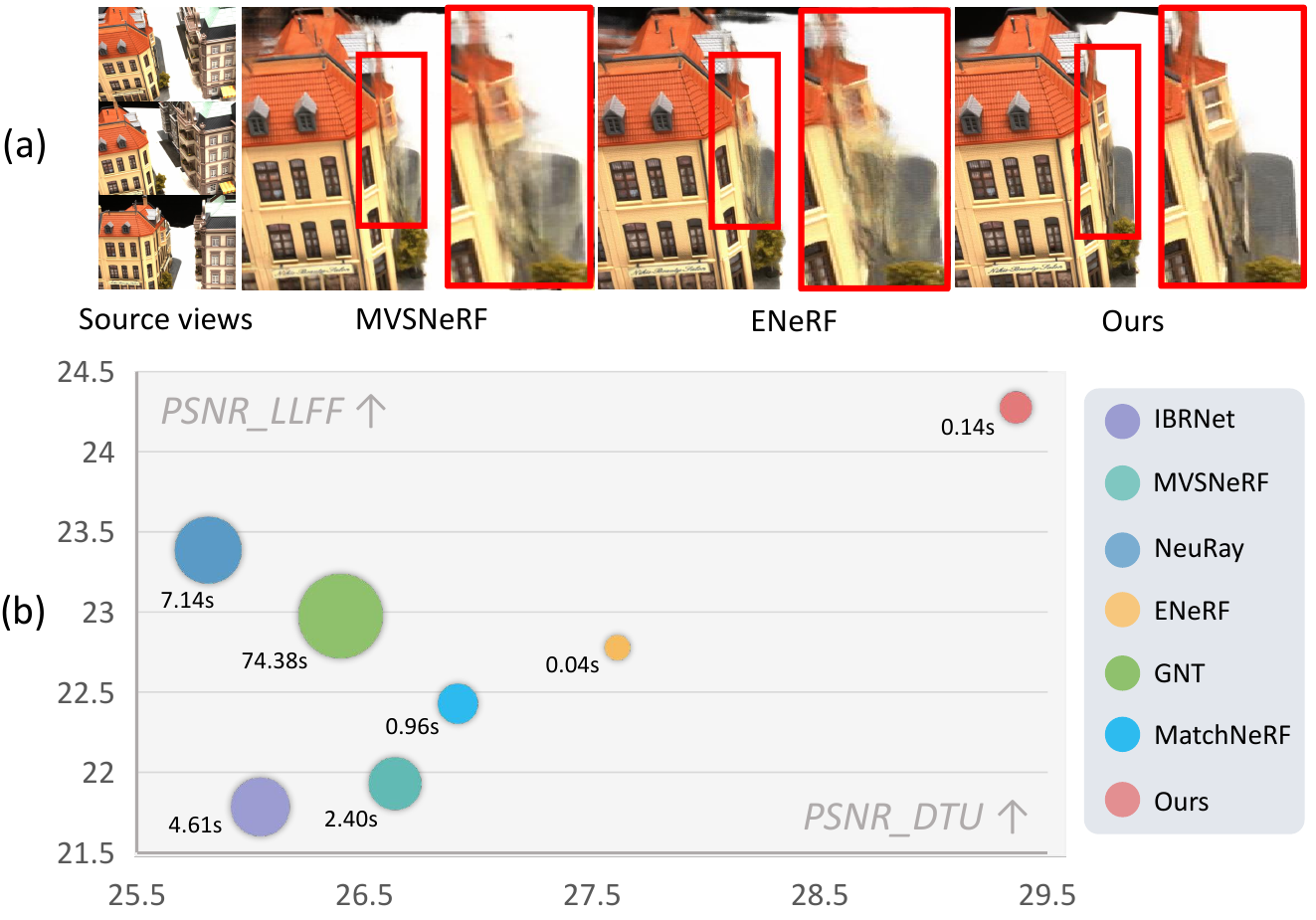}
    \caption{\textbf{Comparison with existing methods.} (a) With three input source views, our generalizable model synthesizes novel views with higher quality than existing methods~\cite{mvsnerf,enerf} in the severe occluded area. (b) Circle area represents inference time. The X-axis represents the PSNR on the DTU dataset~\cite{dtu} and the Y-axis represents the PSNR on the Real Forward-facing dataset~\cite{llff}. Our method attains state-of-the-art performance.}
    \label{fig:fig1}
\end{figure}

\section{Introduction}
\label{sec:introduction}
Novel view synthesis (NVS) aims to generate realistic images at novel viewpoints given a set of posed images. 
By encoding the density and radiance fields of scenes into implicit representations, Neural Radiance Field (NeRF)~\cite{nerf} has shown impressive performance in NVS.
However, NeRF requires a lengthy optimization with densely captured images for each scene, which limits its applications.

To address this issue, some recent methods~\cite{pixelnerf,grf,mine,srf,ibrnet,mvsnerf,pointnerf,neuray,geonerf,enerf,gnt,matchnerf} generalize NeRFs to unseen scenes. 
Instead of overfitting the scene, they extract feature descriptors for 3D points in a scene-agnostic manner, which are then decoded for rendering novel views.
Pioneer methods~\cite{pixelnerf,ibrnet} utilize 2D features warped from the source images.
As this practice avoids explicit modeling of 3D geometric constraints, its generalization capability for geometry reasoning and view synthesis in new scenes is limited.
Hence, subsequent methods~\cite{enerf,mvsnerf} introduce explicit geometry-aware cost volume from multi-view stereo (MVS) to model geometry at the novel view.
Despite significant progress achieved by these methods, synthesis results remain unsatisfactory, especially in challenging areas, such as the occluded area illustrated in Fig.~\ref{fig:fig1} (a).
The model struggles to accurately infer the geometry of these regions, as variance-based cost volumes cannot perceive occlusions.



%
Generalizable NeRFs' pipeline consists of two phases: radiance field reconstruction and rendering.
The reconstruction phase aims to recover scene geometry and encode 3D-aware features for rendering, \ie, creating descriptors for 3D points.
For geometry reasoning, similar to MVS, if the geometry (\ie, depth) is accurate, features across various views are supposed to be similar with low variance.
However, this variance-based cost metric is not universal, especially in occluded and reflective regions.
Due to inconsistent features in these regions, equally considering the contributions of different views is unreasonable, leading to misleading variance values.
Inspired by MVS methods~\cite{aa-rmvsnet,mvster}, we propose an Adaptive Cost Aggregation~(ACA) module.
ACA adaptively reweights the contributions of different views based on the similarity between source views and the novel view.
Since the novel view is unavailable, we employ a coarse-to-fine framework and transfer the rendering view from the coarse stage into the fine stage to learn adaptive weights for the cost volume construction.

With the geometry derived from the cost volume, we can re-sample 3D points around the surface and encode descriptors for sampled points.
For descriptors encoding, previous methods~\cite{pixelnerf,ibrnet,enerf,matchnerf} directly aggregate inter-view features into descriptors for subsequent rendering.
These descriptors lack 3D context awareness, leading to discontinuities in the descriptor space.
To this end, we design the Spatial-View Aggregator~(SVA) to learn 3D context-aware descriptors.
Specifically, we encode spatially context-aware and smooth features by aggregating 3D spatial information.
Meanwhile, to preserve geometric details, we utilize smoothed features as queries to reassemble high-frequency information across views to create final descriptors.
%

\begin{figure}
    \centering
    \includegraphics[width=0.45\textwidth]{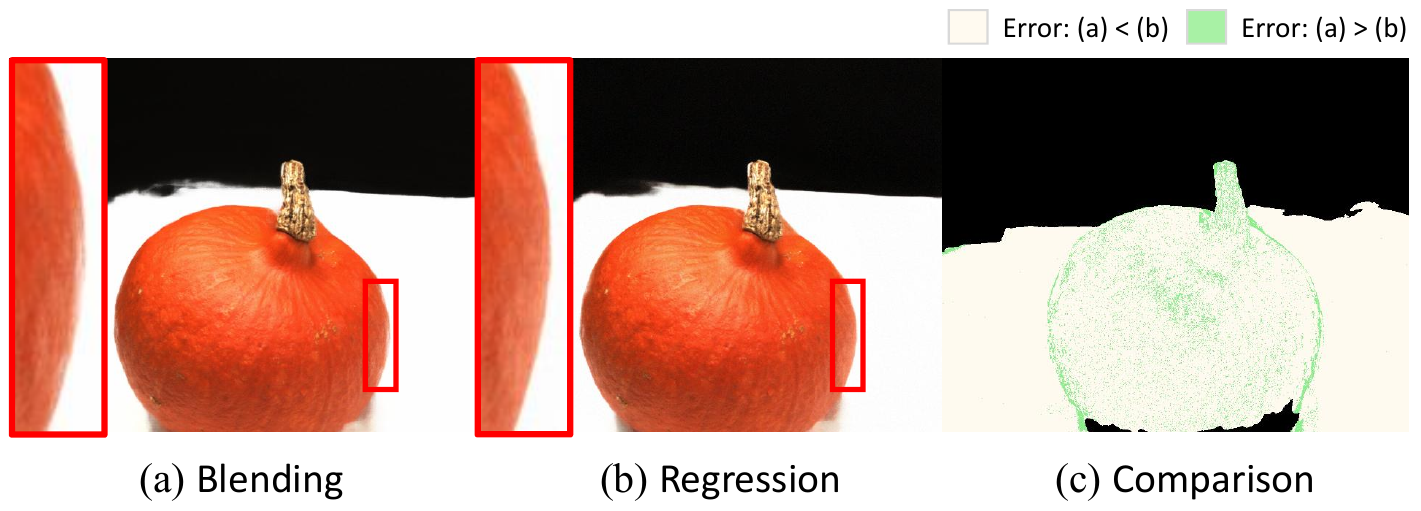}
    \caption{\textbf{Comparison of two rendering strategies.} (a) The view obtained using the blending approach that combines color values from source views. (b) The view obtained using the regression approach that directly regresses color values from features. (c) Accuracy comparison between two rendering strategies. The former strategy performs better in the white regions, while worse in the green ones.}
    \label{fig:motivation}
\end{figure}

%
With the scene geometry and point-wise descriptors, the subsequent rendering phase aims to decode descriptors into volume density and radiance for rendering a novel view.
For the radiance prediction, ~\cite{ibrnet,enerf} predict blending weights to combine color values from source views, while ~\cite{pixelnerf,mvsnerf,matchnerf} directly regress from features.
However, the analysis of these two approaches has not been conducted in existing works.
In this paper, we observe that the blending approach performs better in most areas (Fig.~\ref{fig:motivation} (a)), as the color values from source views provide referential factors.
However, as shown in Fig.~\ref{fig:motivation} (b)$\&$(c), in challenging areas such as reflections and boundaries, the regression approach produces superior results with fewer artifacts, while the blending approach leads to suboptimal rendering due to unreliable referential factors.
To unify the advantages of both strategies, we propose to separately predict two intermediate views using two approaches and design a weighted structure named Consistency-Aware Fusion~(CAF) to dynamically fuse them into the final view.
The fusing weights are learned by checking multi-view consistency, following an underlying principle that if color values are close to the ground truth, the multi-view features corresponding to the correct depth are supposed to be similar.

By embedding the above ACA, SVA, and CAF into a coarse-to-fine framework, we propose \oursnospace.
To demonstrate the effectiveness, we evaluate \ours on the widely-used DTU~\cite{dtu}, Real Forward-facing~\cite{llff}, and NeRF Synthetic~\cite{nerf} datasets.
Extensive experiments show that \ours outperforms other generalizable NeRFs by large margins without scene-specific fine-tuning as shown in Fig.~\ref{fig:fig1} (a)$\&$(b). 
After per-scene fine-tuning, \ours also outperforms other generalizable NeRFs and achieves performance comparable to or even better than NeRF~\cite{nerf}.
Additionally, \ours is capable of generating reasonable depth maps, surpassing other generalization methods.

\noindent Our main contributions can be summarized as follows:
\begin{itemize}
    \item We propose ACA to improve geometry estimation and SVA to encode 3D context-aware descriptors for geometry-aware reconstruction.
    
    \item We conduct an analysis of two existing color decoding strategies and propose CAF to unify their advantages for fusion-refined rendering.

    \item \ours achieves state-of-the-art performance across multiple datasets, showing superior generalization.

\end{itemize}

\section{Related Work}
\label{sec:related work}
\noindent{\textbf{Multi-View Stereo.}}
Given multiple calibrated images, multi-view stereo (MVS) aims to reconstruct a dense 3D representation.
Traditional MVS methods~\cite{fua1995object,gipuma,colmap,schonberger2016pixelwise} primarily rely on hand-crafted features and similarity metrics, which limits their performance, especially in challenging regions such as weak-texture and repetitive areas.
Powered by the impressive representation of neural networks, MVSNet~\cite{mvsnet} first proposes an end-to-end cost volume-based pipeline, which quickly becomes the mainstream in the MVS community.
Following works explore the potential capacity of this pipeline from various aspects.
\eg, reducing memory consumption with recurrent approaches~\cite{aa-rmvsnet,D2HC-RMVSNet,rmvsnet} or coarse-to-fine paradigms~\cite{cheng2020deep,gu2020cascade,yang2020cost,zhang2020visibility}, enhancing feature representations~\cite{transmvsnet,et-mvsnet} and modeling output formats~\cite{unimvs,dmvsnet}.
Another important line is to optimize the cost aggregation~\cite{aa-rmvsnet,mvster} by adaptively weighting contributions from various views.
In this paper, following the spirit, we introduce the adaptive cost aggregation tailored for the NVS task to mitigate the issue of inconsistent features caused by reflections and occlusions.

\noindent \textbf{Generalizable NeRF.}
With implicit continuous representation and differentiable volume rendering, NeRF~\cite{nerf} achieves photo-realistic view synthesis.
However, NeRF and its downstream expansion works~\cite{sofgan,neutex,nerd,bi2020neural,nerfies,xian2021space} require an expensive per-scene optimization process.
To address this issue, some generalizable NeRF methods have been proposed, following a reconstruction-and-rendering pipeline.
In the reconstruction phase, each sampled point is assigned a feature descriptor.
Specifically, according to the descriptors, generalizable NeRF methods can be categorized into the following types: appearance descriptors~\cite{pixelnerf}, aggregated multi-view descriptors~\cite{ibrnet,enerf,gnt}, cost volume interpolated descriptors~\cite{mvsnerf,neuray,enerf}, and correspondence matching descriptors~\cite{matchnerf}.
Despite different forms, these descriptors only aggregate inter-view information or are interpolated from the low-resolution cost volume, lacking the ability to effectively perceive 3D spatial context.
To remedy the issue, we utilize a proposed aggregator to facilitate the interaction of spatial information.
In the rendering phase, volume density is obtained by decoding descriptors. For radiance, ~\cite{ibrnet,enerf} predict blending weights to combine color from source views, while ~\cite{pixelnerf,mvsnerf,matchnerf,gnt,neuray} directly regress features.
In this paper, we observe that these two strategies benefit different regions and thus propose a unified structure to integrate their advantages.

\section{Preliminaries}
\label{sec:preliminaries}
\noindent \textbf{Learning-based MVS.} 
Given a target image and $N$ source images, MVS aims to recover the geometry, such as the depth map of the target image.
The key idea of MVS is to construct the cost volume from multi-view inputs, aggregating 2D information into 3D geometry-aware representation.
Specifically, each voxel-aligned feature vector $f_c$ of cost volume can be computed as:
\begin{equation}
\label{eq:1}
f_c = \Omega (f_t, f_s^1, ..., f_s^N)\,,
\end{equation}
where $f_t$ and $f_s^i$ represent the target feature vector and the warped source feature vector, respectively. And $\Omega$ denotes a consistency metric, such as variance.
The underlying principle is that if a sampled depth is close to the actual depth, the multi-view features of the sampled point are supposed to be similar, which naturally performs multi-view correspondence matching and geometry reasoning, facilitating the generalization to unseen scenes.

\noindent \textbf{Generalizable NeRF.} 
In generalizable NeRFs, each sampled point is assigned a geometry-aware feature descriptor $f_p$, as $(x,d,f_p) \xrightarrow{} (\sigma,r)$, where $x$ and $d$ represent the coordinate and view direction used by NeRF~\cite{nerf}. $\sigma$ and $r$ denote the volume density and radiance for the sampled point, respectively.
Specifically, the volume density $\sigma$ can be obtained from the descriptor via $\sigma = \text{MLP} (x, f_p)$.
For the radiance $r$, one approach~\cite{ibrnet,enerf} is to predict blending weights to combine color values from source views, as:
\begin{equation}\small
\label{eq:2}
\begin{aligned}
r = \sum_{i=1}^{N} \frac{exp(w_i) c_i}{\sum_{j=1}^{N} exp(w_j)} \,,
\text{where }
w_i = \text{MLP} (x,d,f_p, f_s^i) \,,  \\
\end{aligned}
\end{equation}
where $\{c_i\}_{i=1}^N$ are color values from source views.
These colors provide referential factors, facilitating convergence and better performance in most areas.
However, in occluded and reflective regions, these colors introduce misleading bias to the combination, resulting in distorted colors.
Besides, for pixels on object boundaries, it is hard to accurately locate reference points for blending, resulting in oscillated colors between the foreground and background.
Another approach~\cite{mvsnerf,pixelnerf,matchnerf} is to directly regress the radiance from the feature per $r = \text{MLP} (x,d,f_p)$.
As fewer inductive biases are imposed on the output space, the model can learn to predict fewer artifacts in challenging areas (\cref{fig:motivation}). 
%
%
With the volume density and radiance of sampled points, the color values $c$ of each pixel can be computed by volume rendering, given by:
\begin{equation}\small
\label{eq:3}
c = \sum_{k} \tau_k (1-\text{exp}(-\sigma_{k}))r_k \,,
\text{where }
\tau_k = \text{exp}(-\sum_{j=1}^{k-1} \sigma_{j}) \,,
\end{equation}
where $\tau$ represents the volume transmittance.

\begin{figure*}
    \centering
    \includegraphics[width=1.0\textwidth]{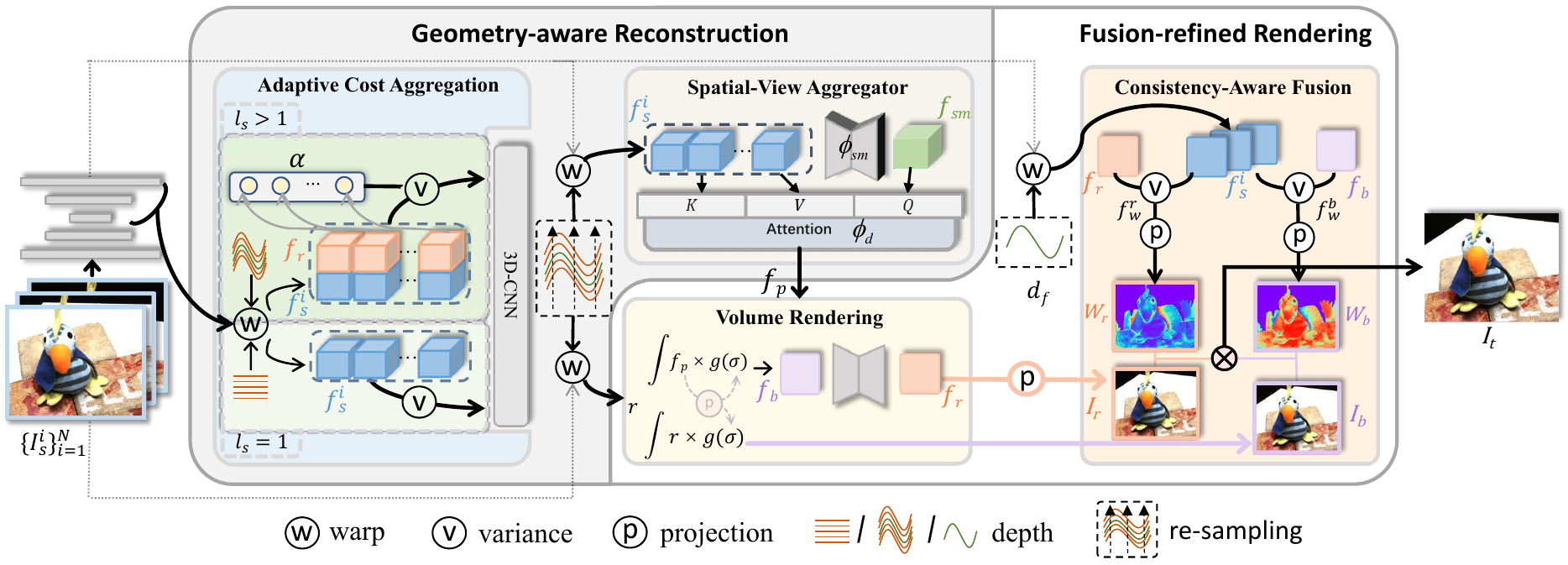}
    \caption{\textbf{The overview of \oursnospace.} In the reconstruction phase, we first infer the geometry from the constructed cost volume, and the geometry guides us to further re-sample 3D points around the surface. For each sampled point, the warped features from source images are aggregated and then fed into our proposed Spatial-View Aggregator (SVA) to learn spatial and inter-view context-aware descriptors $f_p$. In the rendering phase, we apply two decoding strategies to obtain two intermediate views and fuse them into the final target view in an adaptive way, termed Consistency-Aware Fusion (CAF). Our pipeline adopts a coarse-to-fine architecture, the geometry from the coarse stage ($l_s=1$) guides the sampling at the fine stage ($l_s > 1$), and the features from the coarse stage are transferred to the fine stage for ACA to improve geometry estimation. Our network is trained end-to-end using only RGB images.}
    \label{fig:pipeline}
\end{figure*}

\section{Method}
\label{sec:method}
Given a set of source views $\{I_s^i\}_{i=1}^N$, NVS aims to generate a target view at a novel camera pose. 
As illustrated in Fig.~\ref{fig:pipeline}, our method consists of an NVS pipeline wrapped in a coarse-to-fine framework.
%
%
In the pipeline, we first employ a feature pyramid network~\cite{lin2017feature} to extract multi-scale features from the source views.
Then, we propose an Adaptive Cost Aggregation module (\cref{subsec:adaptive cost aggregation}) to construct a cost volume, which is further processed by a 3D-CNN to infer the geometry.
Guided by the estimated geometry, we re-sample 3D points around the surface and apply the Spatial-View Aggregator (\cref{subsec:Spatial-View Aggregator}) to encode 3D-aware feature descriptors $f_p$ for sampled points.
Finally, we decode $f_p$ into two intermediate views using two strategies introduced in~\cref{sec:preliminaries} and fuse them into the final target view through Consistency-Aware Fusion (\cref{subsec: consistency-aware fusion}).
This pipeline is iteratively called. Initially, the low-resolution target view is generated and the rough scene geometry is captured. Then, in the subsequent refining stage, high-resolution results with fine-grained geometry are obtained.

\subsection{Adaptive Cost Aggregation}
\label{subsec:adaptive cost aggregation}
The core process of geometry reasoning is to construct a cost volume that encodes the multi-view feature consistency.
Previous works~\cite{mvsnerf,enerf} treat different views equally and employ the variance operator to construct a cost volume. 
However, due to potential occlusions and varying lighting conditions among different views, multi-view features of the same 3D point may exhibit notable disparities, thereby resulting in misleading variance values.
Inspired by~\cite{aa-rmvsnet}, we propose to adaptively weight the contribution of different views to the cost volume, termed Adaptive Cost Aggregation (ACA).
ACA will suppress the cost contribution from features that are inconsistent with the novel views caused by reflections or occlusions, and enhance the contribution of better-matched pixel pairs.
The voxel-aligned feature $f_c$ of cost volume can be computed as:
\begin{equation}\small
\label{eq:4}
f_c = \frac{1}{N} \sum_{i=1}^{N}(1+\alpha(f_c^i)) \odot f_c^i \,,
\text{where }
f_c^i = (f_s^i - f_t) ^ 2 \,,
\end{equation}
where $f_t$ and $f_s^i$ denote the target feature vector and the warped feature vector of source image $I_s^i$, respectively. $\odot$ denotes Hadamard multiplication and $\alpha(.)$ represents the adaptive weight for each view.
However, an important challenge is that the target view is available in MVS, but not for NVS.
To remedy this, we adopt a coarse-to-fine framework where a coarse novel view is first generated and serves as the target view in \cref{eq:4}.
Specifically, we obtain the coarse-stage feature $f_b$ by accumulating descriptors $f_p$ of sampled points along the ray, as: 
\begin{equation}
    f_b = \sum_{k} \tau_k (1-\text{exp}(-\sigma_{k}))f_{p}^{k}\,.
    \label{eq:another branch}
\end{equation}
The coarse-stage features are then fed into a 2D U-Net for spatial aggregation to obtain $f_r$.
We replace $f_t$ in \cref{eq:4} with the feature $f_r$ to construct a robust cost volume, benefiting the geometry estimation.

\subsection{Spatial-View Aggregator}
\label{subsec:Spatial-View Aggregator}
%
With the estimated geometry, 3D points around objects' surfaces can be re-sampled, and the subsequent step is encoding descriptors for these sampled points.
Existing methods~\cite{ibrnet,enerf} aggregate inter-view features with a pooling network $\rho$ to construct the descriptors $f_p = \rho(\{f_s^i\}_{i=1}^N)$.
However, these descriptors only encode multi-view information and lack the awareness of 3D spatial context, leading to discontinuities in the descriptor space.
To introduce 3D spatial information, a feasible approach is to interpolate the low-resolution regularized cost volume, but it lacks fine-grained details.
To address these issues, we design a 3D-aware descriptors encoding approach.
We first utilize a 3D U-Net termed $\phi_{sm}$ to aggregate 3D spatial context, as:
\begin{equation}
    f_{sm}=\phi_{sm}(\rho(\{f_s^i\}_{i=1}^N))\,.
\end{equation}
However, this may lead to the smoothing of 3D features and cause the loss of some high-frequency geometric details.
Therefore, we use the smoothed features as queries to re-gather inter-view high-frequency details, as:
\begin{equation}
    f_p=\phi_d(f_{sm},\{f_s^i\}_{i=1}^N)\,,
\end{equation}
where $\phi_{d}$ is an attention module. $f_{sm}$ is the input query $q$, and $\{f_s^i\}_{i=1}^N$ are the key sequences $k$ and value sequences $v$. 
The sequence lengths of $q$, $k$, and $v$ are $1$, $N$, and $N$, respectively, which results in only a slight increase in computational costs.

\subsection{Consistency-Aware Fusion}
\label{subsec: consistency-aware fusion}
With the descriptor $f_p$, the volume density $\sigma$ is first acquired through an MLP, and color values can be obtained using two decoding approaches discussed in~\cref{sec:preliminaries}.
We observe that these two approaches exhibit advantages in different areas. To combine their strengths, we propose to predict two intermediate views using these two approaches separately, and then fuse them into the final target view.

Specifically, for the blending approach, the radiance $r$ of each sampled point can be computed using~\cref{eq:2}.
And then the pixel color $c_b$ and feature $f_b$ are obtained via the volume rendering manner per~\cref{eq:3} and~\cref{eq:another branch}, respectively. 
For another regression approach, one practice is predicting radiance for sampled points and then accumulating it into pixel color. In contrast, to reduce computational costs, we accumulate point-wise descriptors into features and then decode them into pixel color. Specifically, we feed the accumulated feature into a 2D U-Net for spatial enhancement to obtain pixel feature $f_r$, followed by an MLP to yield the pixel color $c_r$.

Since these two approaches excel in different areas, instead of using a fixed operator, such as average, we propose dynamically fusing them via $c_t = w_b c_b + w_r c_r$, where $w_b$ and $w_r$ are predicted fusing weights.
As the target view is unavailable, comparing the quality of $c_b$ and $c_r$ becomes a chicken-and-egg problem.
A naive practice is to directly predict fusing weights from features, which is under-constrained~(\cref{subsec:ablations and analysis}).
In contrast, we propose to learn fusing weights by using the multi-view feature consistency as a hint. The underlying motivation is that if the predicted colors closely resemble the ground-truth colors, the corresponding features of the predicted view and source views under the correct depth are supposed to be similar.

Specifically, we first obtain the final predicted depth $d_f$ in a volume rendering-like way per $d_f = \sum_{k} \tau_k (1-\text{exp}(-\sigma_{k}))d_k$, where $d_k$ represents the depth of sampled point. 
With the depth $d_f$, we can obtain the warped features $\{f_s^i\}_{i=1}^N$ from source views, and the multi-view consistency can be computed using variance:
\begin{equation}
\label{eq:11}
f_w^{[b,r]} = var(f_{[b,r]},f_s^1,...,f_s^N) \,.
\end{equation}
We then feed the consistency $f_w^{[b,r]}$ into an MLP to obtain the fusing weight $w_{[b,r]}$, followed by a softmax operator for normalization.
The final target view can be represented in matrix form as $I_t = W_b I_b + W_r I_r$.

\subsection{Loss Function}
Our model is trained end-to-end only using the RGB image as supervision.
Following~\cite{enerf}, we use the mean squared error loss as:
\begin{equation}
\label{eq:mse}
L_{mse} = \frac{1}{N_p}\sum_{i=1}^{N_p}||\hat{c}_i-c_i||^2_2 \,,
\end{equation}
where $N_p$ is the number of pixels and $\hat{c}_i$ and $c_i$ are the ground-truth and predicted pixel color, respectively. 
In addition, the perceptual loss~\cite{lpips} and ssim loss~\cite{ssim} can be applied, as: 
\begin{equation}
\label{eq:lpips_ssim}
\begin{aligned}
L_{perc} &= ||h(\hat{I})-h(I)||, \\
L_{ssim} &= 1-ssim(\hat{I},I) \,,
\end{aligned}
\end{equation}
where $h$ is the perceptual function (a VGG$16$ network). $\hat{I}$ and $I$ are the ground-truth and predicted image patches, respectively. 
The loss at the $k^{th}$ stage is as follows:
\begin{equation}
\label{eq:k_th_loss}
L^{k}=L_{mse}+\lambda_p L_{perc} +\lambda_s L_{ssim} \,,
\end{equation}
where $\lambda_p$ and $\lambda_s$ refer to loss weights.
The overall loss is:
\begin{equation}
\label{eq:overall_loss}
L = \sum_{k=1}^{N_s} \lambda ^k  L^k
\end{equation}
where $N_s$ refers to the number of coarse-to-fine stages and $\lambda ^k$ represents the loss weight of the $k^{th}$ stage.

\begin{table*}
\centering
\setlength{\tabcolsep}{5pt}
\resizebox{0.95\linewidth}{!}{
\begin{tabular}{@{}lccccccccccc@{}}
\toprule
\multirow{2}{*}{Method} & \multirow{2}{*}{Settings} & \multicolumn{3}{c}{DTU~\cite{dtu}} & \multicolumn{3}{c}{Real Forward-facing~\cite{llff}} & \multicolumn{3}{c}{NeRF Synthetic~\cite{nerf}}\\ 
\cmidrule(lr){3-5}\cmidrule(lr){6-8}\cmidrule(lr){9-11}
 & & PSNR $\uparrow$ & SSIM $\uparrow$ & LPIPS $\downarrow$ & PSNR $\uparrow$ & SSIM $\uparrow$ & LPIPS $\downarrow$ & PSNR $\uparrow$ & SSIM $\uparrow$ & LPIPS $\downarrow$ \\
\midrule
PixelNeRF~\cite{pixelnerf} &\multirow{8}{*}{3-view} & 19.31 & 0.789 & 0.382 & 11.24 & 0.486 & 0.671 & 7.39 & 0.658 & 0.411  \\
IBRNet~\cite{ibrnet} & & 26.04 & 0.917 & 0.191 & 21.79 & 0.786 & 0.279 & 22.44 & 0.874 & 0.195  \\ 
MVSNeRF~\cite{mvsnerf}  & & 26.63 & 0.931 & 0.168 & 21.93 & 0.795 & 0.252 & 23.62 & 0.897 & 0.176  \\
NeuRay~\cite{neuray}  & & 25.81 & 0.868 & 0.160 & 23.39 & 0.744 & 0.217 & 24.58 & 0.892 & 0.163 \\
ENeRF$^\dag$~\cite{enerf}  & & 27.61 & 0.956 & 0.091 & 22.78 & 0.808 & 0.209 & 26.65 & 0.947 & 0.072  \\ 
ENeRF~\cite{enerf}  & & 27.61 & 0.957 & 0.089 & 23.63 & 0.843 & 0.182 & 26.17 & 0.943 & 0.085  \\
GNT~\cite{gnt} & & 26.39 & 0.923 & 0.156 & 22.98 & 0.761 & 0.221 & 25.80 & 0.905 & 0.104 \\
MatchNeRF~\cite{matchnerf} & & 26.91 & 0.934 & 0.159 &  22.43 & 0.805 & 0.244 & 23.20 & 0.897 & 0.164  \\
Ours & & \textbf{29.36} & \textbf{0.969} & \textbf{0.064} & \textbf{24.28} & \textbf{0.863} & \textbf{0.162} & \textbf{26.99} & \textbf{0.952} & \textbf{0.070}  \\
\midrule
MVSNeRF~\cite{mvsnerf} & \multirow{6}{*}{2-view} & 24.03 & 0.914 & 0.192 & 20.22 & 0.763 & 0.287 & 20.56 & 0.856 & 0.243 \\
NeuRay~\cite{neuray} & & 24.51 & 0.825 & 0.203 & 22.73 & 0.720 & 0.236 & 22.42 & 0.865 & 0.228  \\ 
ENeRF~\cite{enerf} & & 25.48 & 0.942 & 0.107 & 22.78 & 0.821 & 0.191 & 24.83 & 0.931 & 0.117 \\
GNT~\cite{gnt} & & 24.32 & 0.903 & 0.201 & 20.91 & 0.683 & 0.293 & 23.47 & 0.877 & 0.151 \\
MatchNeRF~\cite{matchnerf} & & 25.03 & 0.919 & 0.181 & 20.59 & 0.775 & 0.276 & 20.57 & 0.864 & 0.200  \\
Ours &  & \textbf{26.98} & \textbf{0.955} & \textbf{0.081} & \textbf{23.39} & \textbf{0.839} & \textbf{0.176} & \textbf{25.30} & \textbf{0.939} & \textbf{0.082} \\
\bottomrule
\end{tabular}}
\caption{\textbf{Quantitative results under the generalization setting.} We show the average results of PSNRs, SSIMs, and LPIPSs on three datasets under two settings for the number of input views. ENeRF$^\dag$ represents results borrowed from the original paper. The comparison methods are organized based on the year of publication.}
\label{Tab:generalization}
\end{table*}

\begin{table*}
\centering
\setlength{\tabcolsep}{5pt}
\resizebox{0.95\linewidth}{!}{
\begin{tabular}{@{}lccccccccc@{}}
\toprule
\multirow{2}{*}{Method} & \multicolumn{3}{c}{DTU~\cite{dtu}} & \multicolumn{3}{c}{Real Forward-facing~\cite{llff}} & \multicolumn{3}{c}{NeRF Synthetic~\cite{nerf}}\\ 
\cmidrule(lr){2-4}\cmidrule(lr){5-7}\cmidrule(lr){8-10}
 & PSNR $\uparrow$ & SSIM $\uparrow$ & LPIPS $\downarrow$ & PSNR $\uparrow$ & SSIM $\uparrow$ & LPIPS $\downarrow$ & PSNR $\uparrow$ & SSIM $\uparrow$ & LPIPS $\downarrow$ \\
\midrule
NeRF$_{10.2h}$~\cite{nerf} & 27.01 & 0.902 & 0.263 & 25.97 & 0.870 & 0.236 & \textbf{30.63} & \textbf{0.962} & 0.093  \\
IBRNet$_{ft-1.0h}$~\cite{ibrnet} & \textbf{31.35} & 0.956 & 0.131 & 24.88 & 0.861 & 0.189 & 25.62 & 0.939 & 0.111  \\ 
MVSNeRF$_{ft-15min}$~\cite{mvsnerf} & 28.51 & 0.933 & 0.179 & 25.45 & 0.877 & 0.192 & 27.07 & 0.931 & 0.168 \\
NeuRay$_{ft-1.0h}$~\cite{neuray} & 26.96 & 0.847 & 0.174 & 24.34 & 0.781 & 0.205 & 25.91 & 0.896 & 0.115 \\
ENeRF$_{ft-1.0h}$~\cite{enerf} & 28.87 & \underline{0.957} & 0.090 & 24.89 & 0.865 & 0.159 & 27.57 & 0.954 & 0.063  \\
Ours$_{ft-15min}$ & 30.10 & \textbf{0.966} & \underline{0.069} & \underline{26.62} & \underline{0.903} & \underline{0.110} & 28.57 & 0.958 & \underline{0.060} \\
Ours$_{ft-1.0h}$ & \underline{30.18} & \textbf{0.966} & \textbf{0.068} & \textbf{26.76} & \textbf{0.905} & \textbf{0.106} & \underline{28.81} & \underline{0.960} & \textbf{0.058}  \\
\bottomrule
\end{tabular}}
\caption{\textbf{Quantitative results under the per-scene optimization setting.} The best result is in \textbf{bold}, and the second-best is \underline{underlined}.}
\label{Tab:fine-tuning}
\end{table*}

\section{Experiments}
\label{sec:experiments}

\subsection{Settings}
\label{subsec:settings}
\noindent \textbf{Datasets.}
Following MVSNeRF~\cite{mvsnerf}, we divide the DTU~\cite{dtu} dataset into $88$ training scenes and $16$ test scenes.
We first train our generalizable model on the $88$ training scenes of the DTU dataset and then evaluate the trained model on the $16$ test scenes.
To further demonstrate the generalization capability of our method, we also test the trained model (without any fine-tuning) on $8$ scenes from the Real
Forward-facing~\cite{llff} dataset and $8$ scenes from the NeRF Synthetic~\cite{nerf} dataset, both of which have significant differences in view distribution and scene content compared to the DTU dataset.
The image resolutions of the DTU, the Real
Forward-facing, and the NeRF Synthetic datasets are $512\times640$, $640\times960$, and $800\times800$, respectively.
The quality of synthesized novel views is measured by PSNR, SSIM~\cite{ssim}, and LPIPS~\cite{lpips} metrics.

\noindent \textbf{Baselines.}
We compare our methods with state-of-the-art generalizable NeRF methods~\cite{pixelnerf,ibrnet,mvsnerf,neuray,enerf,gnt,matchnerf}.  
For generalization with three views and per-scene optimization, we follow the same setting as~\cite{mvsnerf,enerf,matchnerf} and borrow the results of~\cite{mvsnerf,matchnerf,pixelnerf,ibrnet,nerf} from~\cite{mvsnerf,matchnerf}. We evaluate~\cite{enerf} using the official code and trained models.
To keep consistent with the same settings for a fair comparison, such as the number of input views, dataset splitting, view selection, and image resolution, we use the released code and trained model of~\cite{neuray} and retrain~\cite{gnt} with the released code, and evaluate them under our test settings.
For generalization with two views, we borrow the results of~\cite{matchnerf,mvsnerf} from~\cite{matchnerf}. For other baselines~\cite{enerf,neuray,gnt}, we evaluate them using the released model~\cite{neuray,enerf} or the retrained model~\cite{gnt}.

\noindent \textbf{Implementation Details.}
Following~\cite{enerf}, the number of coarse-to-fine stages $N_s$ is set to $2$.
In our coarse-to-fine framework, we sample $64$ and $8$ depth planes for the coarse-level and fine-level cost volumes, respectively.
And we sample $8$ and $2$ points per ray for the coarse-level and fine-level view rendering, respectively.
We set $\lambda_p=0.1$ and $\lambda_s=0.1$ in \cref{eq:k_th_loss}, while $\lambda^1=0.5$ and $\lambda^2=1$ in \cref{eq:overall_loss}.
%
%
We train our model on four RTX 3090 GPUs using the Adam~\cite{adam} optimizer.
%
%
Refer to the supplementary material for more implementation and network details.

\begin{figure*}
    \centering
    \includegraphics[width=1\textwidth]{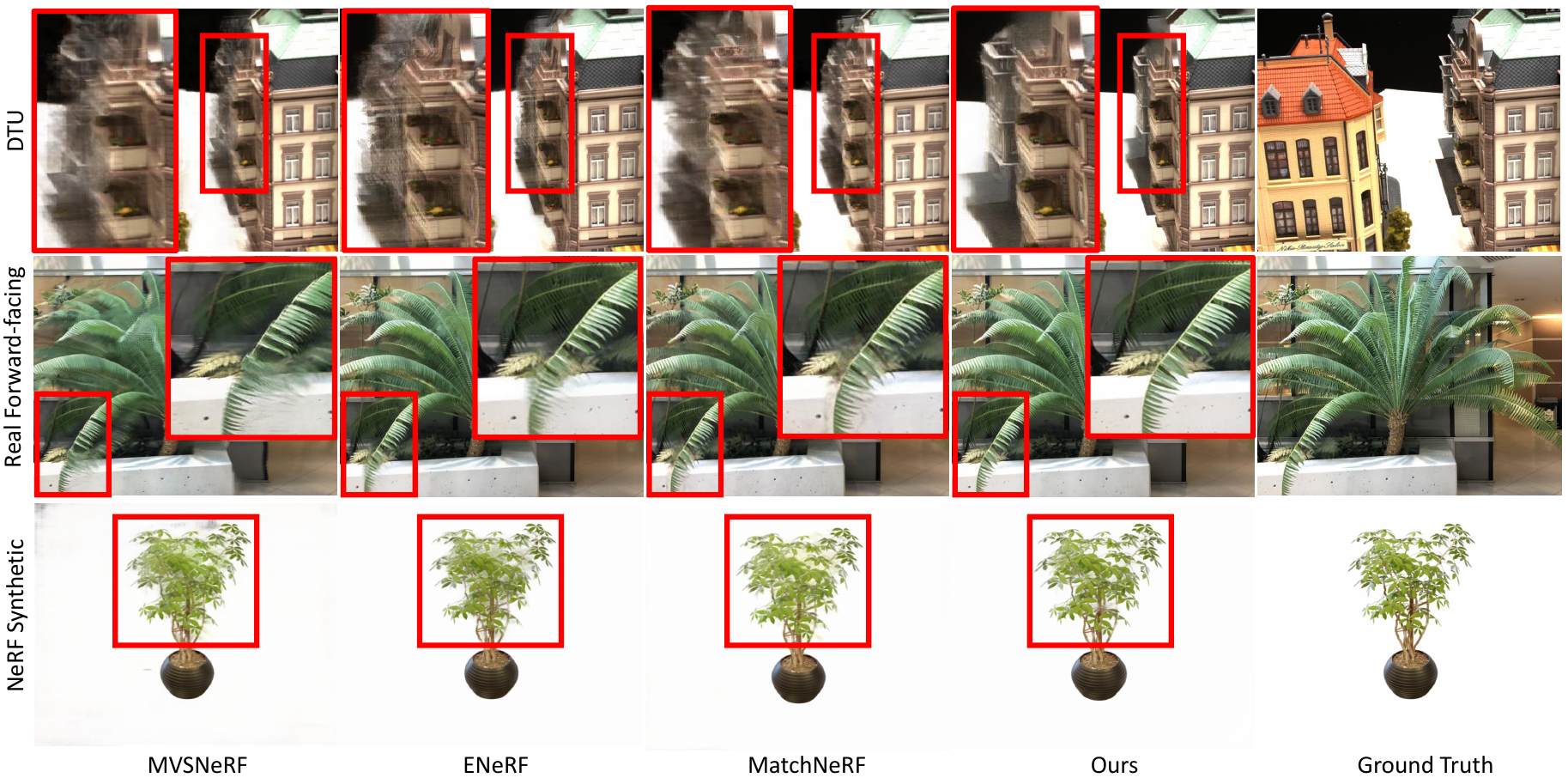}
    \caption{\textbf{Qualitative comparison of rendering quality with state-of-the-art methods~\cite{mvsnerf,enerf,matchnerf} under generalization and three input views settings.}}
    \label{fig:vis_gen}
\end{figure*}

\subsection{Generalization Results}
\label{subsec:generalization results}
We report quantitative results on DTU, Real Forward-facing, and NeRF
Synthetic datasets in Table~\ref{Tab:generalization} under generalization settings.
%
PixelNeRF~\cite{pixelnerf}, which applies appearance descriptors, has reasonable results on the DTU test set, but insufficient generalization on the other two datasets. 
Other methods~\cite{ibrnet,mvsnerf,neuray,enerf,gnt,matchnerf} that model the scene geometry implicitly or explicitly by aggregating multi-view features can maintain relatively good generalization.
Thanks to our proposed modules tailored for both the reconstruction and rendering phases, our method achieves significantly better generalizability.
%
%
As shown in Fig.~\ref{fig:vis_gen}, the views produced by our method preserve more scene details and contain fewer artifacts.
In challenging areas, such as occluded regions, object boundaries, and regions with complex geometry, our method significantly outperforms other methods.
Refer to the supplementary material for more qualitative comparisons.

\begin{figure}
    \centering
    \includegraphics[width=0.42\textwidth]{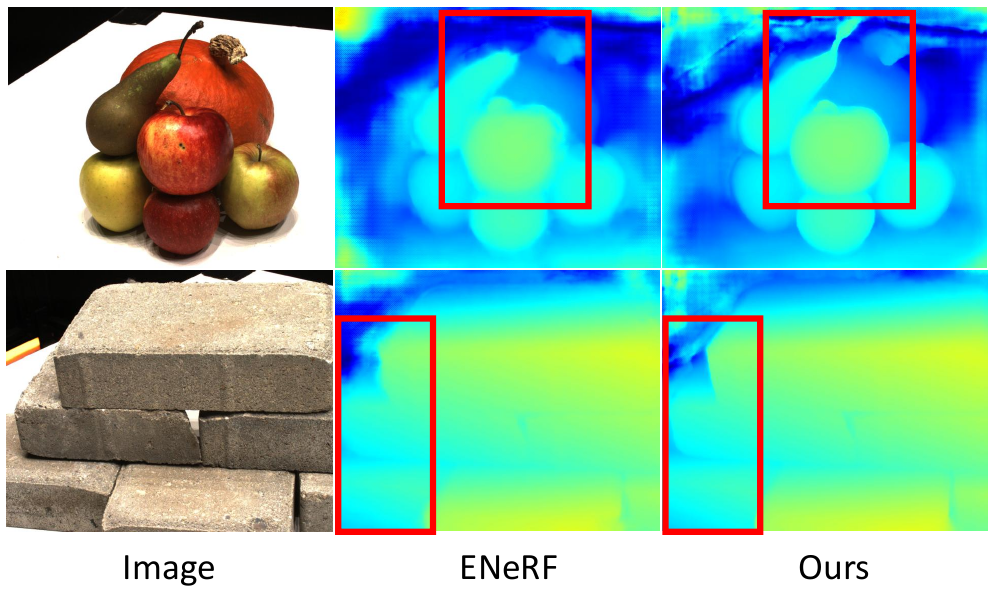}
    \caption{\textbf{Qualitative comparison of depth maps with ~\cite{enerf}.}}
    \label{fig:depth_vis}
\end{figure}

\subsection{Per-scene Fine-tuning Results}
\label{subsec:per-scene fine-tuning results}
The quantitative results after per-scene optimization are shown in Table~\ref{Tab:fine-tuning} and we report the results of our method after $15$ minutes and $1$ hour of fine-tuning.
Due to the excellent initialization provided by our generalization model, only a short period of fine-tuning is needed to achieve good results.
Our results after $15$ minutes of fine-tuning are comparable to or even superior to those of NeRF~\cite{nerf} optimized for substantially longer time ($10.2$ hours), and also outperform the results of other generalization methods after fine-tuning.
With a longer fine-tuning duration, such as $1$ hour, the rendering quality can be further improved.
Qualitative results can be found in the supplementary material.

\subsection{Depth Reconstruction Results}
\label{subsec:depth reconstruction results}
%
Following~\cite{mvsnerf,enerf}, we report the performance of depth reconstruction in Table~\ref{Tab:depth}.
%
%
Our method can achieve higher depth accuracy than other methods, even including the MVS method MVSNet~\cite{mvsnet} that is trained with depth supervision.
As shown in Fig.~\ref{fig:depth_vis}, the depth map produced by our method is more refined, such as sharper object edges.
%

\begin{table}
\centering
\setlength{\tabcolsep}{1pt}
\renewcommand\arraystretch {1}
\scalebox{1}{
\resizebox{\linewidth}{!}{
\begin{tabular}{@{}lcccccc@{}}
\toprule
\multirow{2}{*}{Method} & \multicolumn{3}{c}{Reference view} & \multicolumn{3}{c}{Novel view} \\ 
\cmidrule(lr){2-4}\cmidrule(lr){5-7}
& Abs err $\downarrow$ & Acc(2)$\uparrow$ & Acc(10)$\uparrow$ & Abs err $\downarrow$ & Acc(2)$\uparrow$ & Acc(10)$\uparrow$ \\
\midrule
MVSNet~\cite{mvsnet} & 3.60 & 0.603 & 0.955 & - & - & - \\
PixelNeRF~\cite{pixelnerf} & 49 & 0.037 & 0.176 & 47.8 & 0.039 & 0.187 \\
IBRNet~\cite{ibrnet} & 338 & 0.000 & 0.913 & 324 & 0.000 & 0.866  \\
MVSNeRF~\cite{mvsnerf} & 4.60 & 0.746 & 0.913 & 7.00 & 0.717 & 0.866 \\
ENeRF~\cite{enerf} & 3.80 & 0.837 & 0.939 & 4.60 & 0.792 & 0.917 \\
Ours & \textbf{2.47} & \textbf{0.900} & \textbf{0.971} & \textbf{2.83} & \textbf{0.879} & \textbf{0.961}
\\
\bottomrule
\end{tabular}}
}
\caption{\textbf{Quantitative results of depth reconstruction on the DTU test set.} MVSNet is trained with depth supervision while other methods are trained with only RGB image supervision. ``Abs err'' represents the average absolute error and ``Acc(X)'' means the percentage of pixels with an error less than X mm.}
\label{Tab:depth}
\end{table}

\begin{figure}
    \centering
    \includegraphics[width=0.42\textwidth]{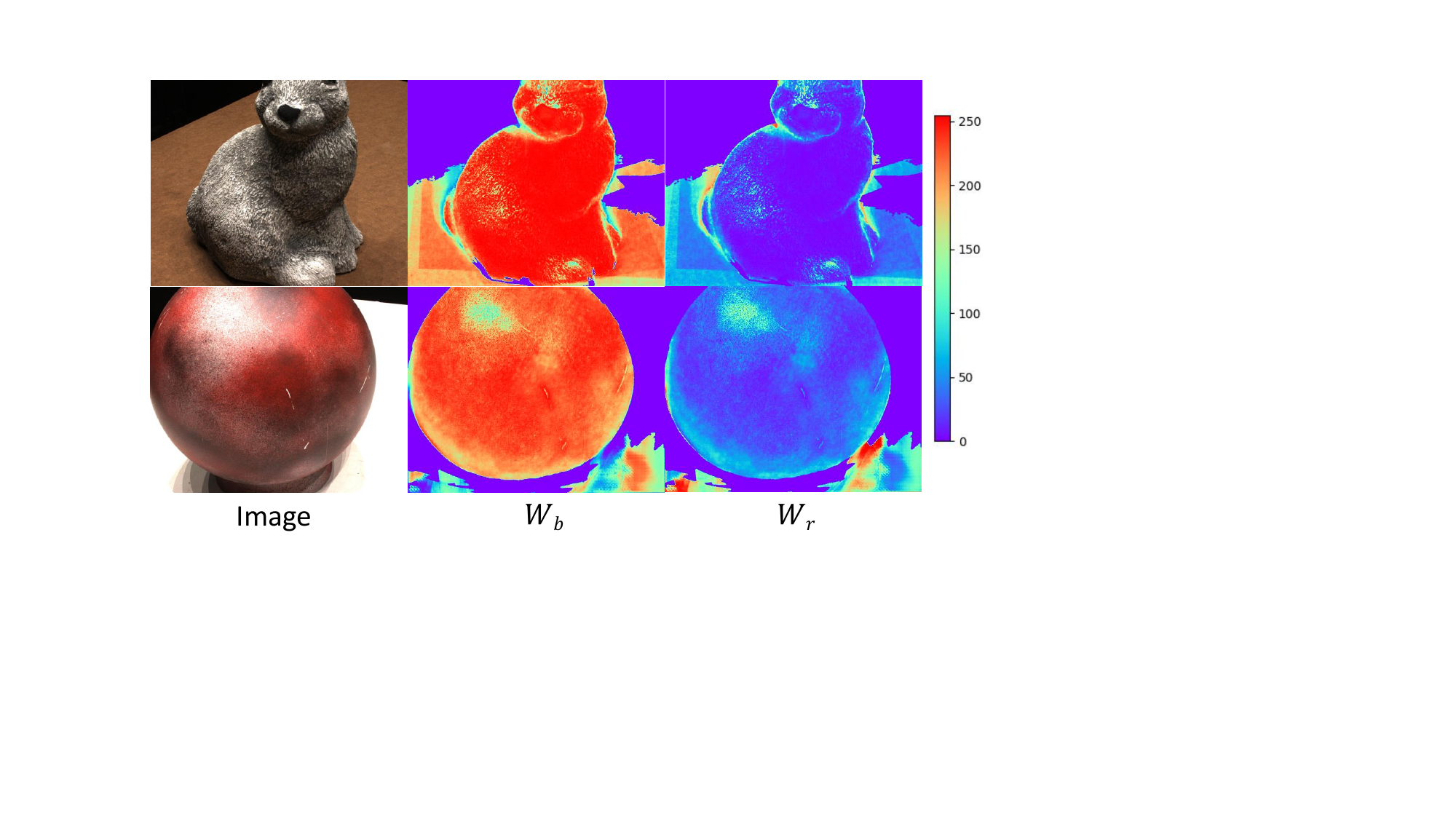}
    \caption{\textbf{Visualization of Fusion Weights.} $W_b$ and $W_r$ represent the weight maps of the blending approach and regression approach, respectively.}
    \label{fig:weight_vis}
\end{figure}

\begin{figure}
    \centering
    \includegraphics[width=0.47\textwidth]{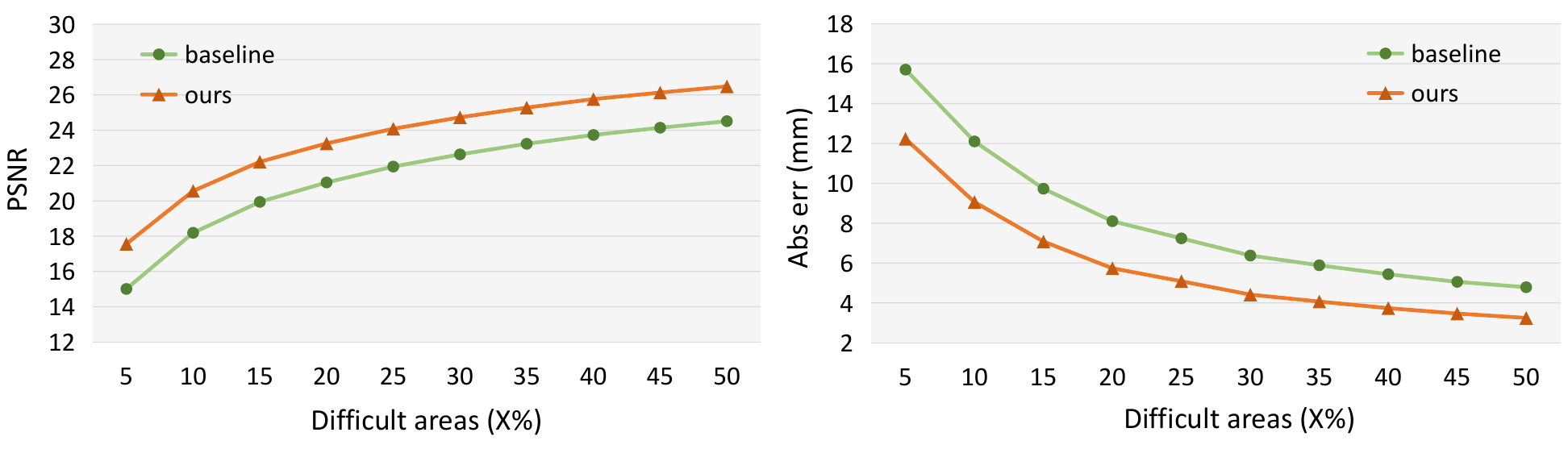}
    \caption{\textbf{Quantitative analysis of difficult areas.} The X-axis difficult areas ($X\%$) represents considering the area with the top $X\%$ of values in $W_r$ as a difficult area. A smaller threshold $X$ indicates a more challenging area.}
    \label{fig:hard_region}
\end{figure}

\subsection{Ablations and Analysis}
\label{subsec:ablations and analysis}

\noindent \textbf{Ablation studies.}
As shown in Table~\ref{Tab:ablation}, we conduct ablation studies to investigate the contribution of each proposed module.
Each individual component can benefit the baseline model in both view quality and depth accuracy, with CAF having the highest gain.
An interesting phenomenon is that CAF greatly improves depth accuracy, indicating that a well-designed view decoding approach also facilitates depth prediction.
Combining all components results in the greatest gain, with a $5.9\%$ increase in PSNR and a $34.5\%$ improvement in depth error compared to the baseline model.

\noindent \textbf{CAF working mechanism.}
As shown in Fig.~\ref{fig:weight_vis}, we visualize the fusion weights of two decoding approaches.
The regression approach exhibits higher confidence in challenging areas such as object edges and reflections, while the blending approach shows higher confidence in most other areas, which is consistent with the observation in Fig.~\ref{fig:motivation}.
As shown in Fig.~\ref{fig:hard_region}, we define challenging areas as those with high confidence in $W_r$ and divide them by a series of thresholds.
A smaller threshold $X$ indicates a more difficult region.
When $X=5\%$, our method improves PSNR by $2.55db$ and ``Abs err'' by $3.47mm$. When the threshold increases, such as $X=50\%$, our method improves PSNR by $1.97db$ and ``Abs err'' by $1.54mm$, which further demonstrates the superiority of our method in challenging areas.

\noindent \textbf{Fusion strategy.}
As shown in Table~\ref{Tab:fusion strategy}, we investigate the performance of different fusion strategies.
Comparing No.1 and No.2, the result of using the blending approach alone is better than that of using the regression approach alone.
%
%
For No.4, the DWF represents fusion weights derived directly from features of the two intermediate views, which greatly degrades performance compared to our proposed way of checking the multi-view consistency (No.5).
Our fusion approach utilizes the advantages of the two decoding approaches to refine the synthesized view. In No.3, we refine the synthesized view decoded in a single way, where the synthesized view is fed into an auto-encoder for refinement, which has limited improvement.

\begin{table}
\centering
\renewcommand\arraystretch {1}
\scalebox{0.9}{
\resizebox{\linewidth}{!}{
\begin{tabular}{@{}cccc|c|c@{}}
\toprule
ACA & SVA$_{sm}$ & SVA$_d$ & CAF & PSNR $\uparrow$  & Abs err $\downarrow$  \\
\midrule
\ding{55} & \ding{55} & \ding{55} & \ding{55} & 27.73 & 4.32 \\
\CheckmarkBold & \ding{55} & \ding{55} & \ding{55} & 28.35 & 3.85 \\
\ding{55} & \CheckmarkBold & \ding{55} & \ding{55} & 28.30 & 3.53 \\
\ding{55} & \ding{55} & \CheckmarkBold & \ding{55} & 27.89 & 3.94 \\
\ding{55} & \ding{55} & \ding{55} & \CheckmarkBold & 28.85 & 3.36 \\
\CheckmarkBold & \CheckmarkBold & \ding{55} & \ding{55} & 28.53 & 3.54 \\
\CheckmarkBold & \CheckmarkBold & \CheckmarkBold & \ding{55} & 28.64 & 3.06\\
\CheckmarkBold & \CheckmarkBold & \CheckmarkBold & \CheckmarkBold & 29.36 & 2.83 \\
\bottomrule
\end{tabular}}}
\caption{\textbf{Ablation studies on the DTU dataset.} We report the image quality~(PSNR) and depth accuracy~(Abs err) metrics without per-scene fine-tuning under three input views settings. SVA$_{sm}$ and SVA$_{d}$ represent $\phi_{sm}$ and $\phi_{d}$ of SVA, respectively.}
\label{Tab:ablation}
\end{table}

\begin{table}
\centering
\setlength{\tabcolsep}{1pt}
\renewcommand\arraystretch {1}
\scalebox{0.9}{
\resizebox{\linewidth}{!}{
\begin{tabular}{@{}c|ccc|ccc@{}}
\toprule
& Blending & Regression & Fusion & PSNR $\uparrow$ & SSIM $\uparrow$ & LPIPS $\downarrow$ \\
\midrule
No.1 & \CheckmarkBold & \ding{55} & \ding{55} & 27.73 & 0.956 & 0.088 \\
No.2 & \ding{55} & \CheckmarkBold & \ding{55} & 26.83 & 0.955 & 0.091 \\
No.3 & \CheckmarkBold & \ding{55} & AE & 27.89 &  0.959 & 0.081 \\
No.4 & \CheckmarkBold & \CheckmarkBold & DWF & 28.10 & 0.963  &  0.075 \\
No.5 & \CheckmarkBold & \CheckmarkBold & CAF & \textbf{28.85} & \textbf{0.966} & \textbf{0.070} \\

\bottomrule
\end{tabular}}
}
\caption{\textbf{Comparison of different fusion strategies.} The AE represents the refinement by an autoencoder. The DWF represents direct weighted fusion. The CAF is our proposed Consistency-Aware Fusion.}
\label{Tab:fusion strategy}
\end{table}

\section{Conclusion}
\label{sec:conclusion}
In this paper, we present a generalizable NeRF method capable of achieving high-fidelity view synthesis.
Specifically, during the reconstruction phase, we propose Adaptive Cost Aggregation (ACA) to improve geometry estimation and Spatial-View Aggregator (SVA) to encode 3D context-aware descriptors.
%
In the rendering phase, we introduce the Consistency-Aware Fusion (CAF) module to unify their advantages to refine the synthesized view quality.
We integrate these modules into a coarse-to-fine framework, termed \oursnospace.
Extensive evaluations and ablations demonstrate the effectiveness of our proposed modules.

{
    \small
    \bibliographystyle{ieeenat_fullname}
    \bibliography{main}

\begin{thebibliography}{46}
\providecommand{\natexlab}[1]{#1}
\providecommand{\url}[1]{\texttt{#1}}
\expandafter\ifx\csname urlstyle\endcsname\relax
  \providecommand{\doi}[1]{doi: #1}\else
  \providecommand{\doi}{doi: \begingroup \urlstyle{rm}\Url}\fi

\bibitem[Aan{ae}s et~al.(2016)Aan{ae}s, Jensen, Vogiatzis, Tola, and Dahl]{dtu}
Henrik Aan{ae}s, Rasmus~Ramsb{o}l Jensen, George Vogiatzis, Engin Tola, and Anders~Bjorholm Dahl.
\newblock Large-scale data for multiple-view stereopsis.
\newblock \emph{Int. J. Comput. Vis.}, 120:\penalty0 153--168, 2016.

\bibitem[Bi et~al.(2020)Bi, Xu, Srinivasan, Mildenhall, Sunkavalli, Ha{\v{s}}an, Hold-Geoffroy, Kriegman, and Ramamoorthi]{bi2020neural}
Sai Bi, Zexiang Xu, Pratul Srinivasan, Ben Mildenhall, Kalyan Sunkavalli, Milo{\v{s}} Ha{\v{s}}an, Yannick Hold-Geoffroy, David Kriegman, and Ravi Ramamoorthi.
\newblock Neural reflectance fields for appearance acquisition.
\newblock \emph{arXiv preprint arXiv:2008.03824}, 2020.

\bibitem[Boss et~al.(2021)Boss, Braun, Jampani, Barron, Liu, and Lensch]{nerd}
Mark Boss, Raphael Braun, Varun Jampani, Jonathan~T Barron, Ce Liu, and Hendrik Lensch.
\newblock Nerd: Neural reflectance decomposition from image collections.
\newblock In \emph{Proc. IEEE Int. Conf. Comput. Vis.}, pages 12684--12694, 2021.

\bibitem[Chang et~al.(2022)Chang, Bo{\v{z}}i{\v{c}}, Zhang, Yan, Chen, S{\"u}sstrunk, and Nie{\ss}ner]{RC-MVSNet}
Di Chang, Alja{\v{z}} Bo{\v{z}}i{\v{c}}, Tong Zhang, Qingsong Yan, Yingcong Chen, Sabine S{\"u}sstrunk, and Matthias Nie{\ss}ner.
\newblock Rc-mvsnet: Unsupervised multi-view stereo with neural rendering.
\newblock In \emph{Proc. Eur. Conf. Comput. Vis.}, 2022.

\bibitem[Chen et~al.(2021)Chen, Xu, Zhao, Zhang, Xiang, Yu, and Su]{mvsnerf}
Anpei Chen, Zexiang Xu, Fuqiang Zhao, Xiaoshuai Zhang, Fanbo Xiang, Jingyi Yu, and Hao Su.
\newblock Mvsnerf: Fast generalizable radiance field reconstruction from multi-view stereo.
\newblock In \emph{Proc. IEEE Int. Conf. Comput. Vis.}, pages 14124--14133, 2021.

\bibitem[Chen et~al.(2022)Chen, Liu, Xie, Chen, Su, and Yu]{sofgan}
Anpei Chen, Ruiyang Liu, Ling Xie, Zhang Chen, Hao Su, and Jingyi Yu.
\newblock Sofgan: A portrait image generator with dynamic styling.
\newblock \emph{ACM Trans. Graph.}, 41\penalty0 (1):\penalty0 1--26, 2022.

\bibitem[Chen et~al.(2023)Chen, Xu, Wu, Zheng, Cham, and Cai]{matchnerf}
Yuedong Chen, Haofei Xu, Qianyi Wu, Chuanxia Zheng, Tat-Jen Cham, and Jianfei Cai.
\newblock Explicit correspondence matching for generalizable neural radiance fields.
\newblock \emph{arXiv preprint arXiv:2304.12294}, 2023.

\bibitem[Cheng et~al.(2020)Cheng, Xu, Zhu, Li, Li, Ramamoorthi, and Su]{cheng2020deep}
Shuo Cheng, Zexiang Xu, Shilin Zhu, Zhuwen Li, Li~Erran Li, Ravi Ramamoorthi, and Hao Su.
\newblock Deep stereo using adaptive thin volume representation with uncertainty awareness.
\newblock In \emph{Proc. IEEE Conf. Comput. Vis. Pattern Recogn.}, pages 2524--2534, 2020.

\bibitem[Chibane et~al.(2021)Chibane, Bansal, Lazova, and Pons-Moll]{srf}
Julian Chibane, Aayush Bansal, Verica Lazova, and Gerard Pons-Moll.
\newblock Stereo radiance fields (srf): Learning view synthesis for sparse views of novel scenes.
\newblock In \emph{Proc. IEEE Conf. Comput. Vis. Pattern Recogn.}, pages 7911--7920, 2021.

\bibitem[Ding et~al.(2022)Ding, Yuan, Zhu, Zhang, Liu, Wang, and Liu]{transmvsnet}
Yikang Ding, Wentao Yuan, Qingtian Zhu, Haotian Zhang, Xiangyue Liu, Yuanjiang Wang, and Xiao Liu.
\newblock Transmvsnet global context-aware multi-view stereo network with transformers.
\newblock In \emph{Proc. IEEE Conf. Comput. Vis. Pattern Recogn.}, pages 8585--8594, 2022.

\bibitem[Fua and Leclerc(1995)]{fua1995object}
Pascal Fua and Yvan~G Leclerc.
\newblock Object-centered surface reconstruction combining multi-image stereo and shading.
\newblock \emph{Int. J. Comput. Vis.}, 16\penalty0 (ARTICLE):\penalty0 35--56, 1995.

\bibitem[Galliani et~al.(2015)Galliani, Lasinger, and Schindler]{gipuma}
Silvano Galliani, Katrin Lasinger, and Konrad Schindler.
\newblock Massively parallel multiview stereopsis by surface normal diffusion.
\newblock In \emph{Proc. IEEE Int. Conf. Comput. Vis.}, pages 873--881, 2015.

\bibitem[Girshick(2015)]{FastR-CNN}
Ross Girshick.
\newblock Fast r-cnn.
\newblock In \emph{Proc. IEEE Int. Conf. Comput. Vis.}, 2015.

\bibitem[Gu et~al.(2020)Gu, Fan, Zhu, Dai, Tan, and Tan]{gu2020cascade}
Xiaodong Gu, Zhiwen Fan, Siyu Zhu, Zuozhuo Dai, Feitong Tan, and Ping Tan.
\newblock Cascade cost volume for high-resolution multi-view stereo and stereo matching.
\newblock In \emph{Proc. IEEE Conf. Comput. Vis. Pattern Recogn.}, pages 2495--2504, 2020.

\bibitem[Johari et~al.(2022)Johari, Lepoittevin, and Fleuret]{geonerf}
M. Johari, Y. Lepoittevin, and F. Fleuret.
\newblock Geonerf: Generalizing nerf with geometry priors.
\newblock In \emph{Proc. IEEE Conf. Comput. Vis. Pattern Recogn.}, 2022.

\bibitem[Kerbl et~al.(2023)Kerbl, Kopanas, Leimk{\"u}hler, and Drettakis]{3Dgaussians}
Bernhard Kerbl, Georgios Kopanas, Thomas Leimk{\"u}hler, and George Drettakis.
\newblock 3d gaussian splatting for real-time radiance field rendering.
\newblock \emph{ACM Transactions on Graphics}, 42\penalty0 (4), 2023.

\bibitem[Khot et~al.(2019)Khot, Agrawal, Tulsiani, Mertz, Lucey, and Hebert]{khot2019learning}
Tejas Khot, Shubham Agrawal, Shubham Tulsiani, Christoph Mertz, Simon Lucey, and Martial Hebert.
\newblock Learning unsupervised multi-view stereopsis via robust photometric consistency.
\newblock \emph{arXiv preprint arXiv:1905.02706}, 2019.

\bibitem[Kingma and Ba(2014)]{adam}
Diederik~P Kingma and Jimmy Ba.
\newblock Adam: A method for stochastic optimization.
\newblock \emph{arXiv preprint arXiv:1412.6980}, 2014.

\bibitem[Li et~al.(2021)Li, Feng, She, Ding, Wang, and Lee]{mine}
Jiaxin Li, Zijian Feng, Qi She, Henghui Ding, Changhu Wang, and Gim~Hee Lee.
\newblock Mine: Towards continuous depth mpi with nerf for novel view synthesis.
\newblock In \emph{Proc. IEEE Int. Conf. Comput. Vis.}, 2021.

\bibitem[Lin et~al.(2022)Lin, Peng, Xu, Yan, Shuai, Bao, and Zhou]{enerf}
Haotong Lin, Sida Peng, Zhen Xu, Yunzhi Yan, Qing Shuai, Hujun Bao, and Xiaowei Zhou.
\newblock Efficient neural radiance fields for interactive free-viewpoint video.
\newblock In \emph{SIGGRAPH Asia Conference Proceedings}, 2022.

\bibitem[Lin et~al.(2017)Lin, Doll{\'a}r, Girshick, He, Hariharan, and Belongie]{lin2017feature}
Tsung-Yi Lin, Piotr Doll{\'a}r, Ross Girshick, Kaiming He, Bharath Hariharan, and Serge Belongie.
\newblock Feature pyramid networks for object detection.
\newblock In \emph{Proceedings of the IEEE conference on computer vision and pattern recognition}, pages 2117--2125, 2017.

\bibitem[Liu et~al.(2023)Liu, Ye, Zhao, Pan, Shi, and Cao]{et-mvsnet}
Tianqi Liu, Xinyi Ye, Weiyue Zhao, Zhiyu Pan, Min Shi, and Zhiguo Cao.
\newblock When epipolar constraint meets non-local operators in multi-view stereo.
\newblock In \emph{Proc. IEEE Int. Conf. Comput. Vis.}, pages 18088--18097, 2023.

\bibitem[Liu et~al.(2022)Liu, Peng, Liu, Wang, Wang, Theobalt, Zhou, and Wang]{neuray}
Yuan Liu, Sida Peng, Lingjie Liu, Qianqian Wang, Peng Wang, Christian Theobalt, Xiaowei Zhou, and Wenping Wang.
\newblock Neural rays for occlusion-aware image-based rendering.
\newblock In \emph{Proc. IEEE Conf. Comput. Vis. Pattern Recogn.}, 2022.

\bibitem[Mildenhall et~al.(2019)Mildenhall, Srinivasan, Ortiz-Cayon, Kalantari, Ramamoorthi, Ng, and Kar]{llff}
Ben Mildenhall, Pratul~P Srinivasan, Rodrigo Ortiz-Cayon, Nima~Khademi Kalantari, Ravi Ramamoorthi, Ren Ng, and Abhishek Kar.
\newblock Local light field fusion: Practical view synthesis with prescriptive sampling guidelines.
\newblock \emph{ACM Trans. Graph.}, 38\penalty0 (4):\penalty0 1--14, 2019.

\bibitem[Mildenhall et~al.(2020)Mildenhall, Srinivasan, Tancik, Barron, Ramamoorthi, and Ng]{nerf}
Ben Mildenhall, Pratul~P. Srinivasan, Matthew Tancik, Jonathan~T. Barron, Ravi Ramamoorthi, and Ren Ng.
\newblock Nerf: Representing scenes as neural radiance fields for view synthesis.
\newblock In \emph{Proc. Eur. Conf. Comput. Vis.}, 2020.

\bibitem[Park et~al.(2021)Park, Sinha, Barron, Bouaziz, Goldman, Seitz, and Martin-Brualla]{nerfies}
Keunhong Park, Utkarsh Sinha, Jonathan~T Barron, Sofien Bouaziz, Dan~B Goldman, Steven~M Seitz, and Ricardo Martin-Brualla.
\newblock Nerfies: Deformable neural radiance fields.
\newblock In \emph{Proc. IEEE Int. Conf. Comput. Vis.}, pages 5865--5874, 2021.

\bibitem[Peng et~al.(2022)Peng, Wang, Wang, Lai, and Wang]{unimvs}
Rui Peng, Rongjie Wang, Zhenyu Wang, Yawen Lai, and Ronggang Wang.
\newblock Rethinking depth estimation for multi-view stereo a unified representation.
\newblock In \emph{Proc. IEEE Conf. Comput. Vis. Pattern Recogn.}, pages 8645--8654, 2022.

\bibitem[Schonberger and Frahm(2016)]{colmap}
Johannes~L Schonberger and Jan-Michael Frahm.
\newblock Structure-from-motion revisited.
\newblock In \emph{Proc. IEEE Conf. Comput. Vis. Pattern Recogn.}, pages 4104--4113, 2016.

\bibitem[Sch{o}nberger et~al.(2016)Sch{o}nberger, Zheng, Frahm, and Pollefeys]{schonberger2016pixelwise}
Johannes~L Sch{o}nberger, Enliang Zheng, Jan-Michael Frahm, and Marc Pollefeys.
\newblock Pixelwise view selection for unstructured multi-view stereo.
\newblock In \emph{Proc. Eur. Conf. Comput. Vis.}, pages 501--518. Springer, 2016.

\bibitem[T et~al.(2023)T, Wang, Chen, Chen, Venugopalan, and Wang]{gnt}
Mukund~Varma T, Peihao Wang, Xuxi Chen, Tianlong Chen, Subhashini Venugopalan, and Zhangyang Wang.
\newblock Is attention all that ne{RF} needs?
\newblock In \emph{Proc. Int. Conf. Learn. Repr.}, 2023.

\bibitem[Trevithick and Yang(2021)]{grf}
Alex Trevithick and Bo Yang.
\newblock Grf: Learning a general radiance field for 3d representation and rendering.
\newblock In \emph{Proc. IEEE Int. Conf. Comput. Vis.}, pages 15182--15192, 2021.

\bibitem[Wang et~al.(2021)Wang, Wang, Genova, Srinivasan, Zhou, Barron, Martin-Brualla, Snavely, and Funkhouser]{ibrnet}
Qianqian Wang, Zhicheng Wang, Kyle Genova, Pratul Srinivasan, Howard Zhou, Jonathan~T. Barron, Ricardo Martin-Brualla, Noah Snavely, and Thomas Funkhouser.
\newblock Ibrnet: Learning multi-view image-based rendering.
\newblock In \emph{Proc. IEEE Conf. Comput. Vis. Pattern Recogn.}, 2021.

\bibitem[Wang et~al.(2022)Wang, Zhu, Huang, Qin, Ye, He, Chi, and Wang]{mvster}
Xiaofeng Wang, Zheng Zhu, Guan Huang, Fangbo Qin, Yun Ye, Yijia He, Xu Chi, and Xingang Wang.
\newblock Mvster epipolar transformer for efficient multi-view stereo.
\newblock In \emph{Proc. Eur. Conf. Comput. Vis.}, pages 573--591. Springer, 2022.

\bibitem[Wang et~al.(2004)Wang, Bovik, Sheikh, and Simoncelli]{ssim}
Zhou Wang, Alan~C Bovik, Hamid~R Sheikh, and Eero~P Simoncelli.
\newblock Image quality assessment: from error visibility to structural similarity.
\newblock \emph{IEEE transactions on image processing}, 13\penalty0 (4):\penalty0 600--612, 2004.

\bibitem[Wei et~al.(2021)Wei, Zhu, Min, Chen, and Wang]{aa-rmvsnet}
Zizhuang Wei, Qingtian Zhu, Chen Min, Yisong Chen, and Guoping Wang.
\newblock Aa-rmvsnet adaptive aggregation recurrent multi-view stereo network.
\newblock In \emph{Proc. IEEE Int. Conf. Comput. Vis.}, pages 6187--6196, 2021.

\bibitem[Xian et~al.(2021)Xian, Huang, Kopf, and Kim]{xian2021space}
Wenqi Xian, Jia-Bin Huang, Johannes Kopf, and Changil Kim.
\newblock Space-time neural irradiance fields for free-viewpoint video.
\newblock In \emph{Proc. IEEE Conf. Comput. Vis. Pattern Recogn.}, pages 9421--9431, 2021.

\bibitem[Xiang et~al.(2021)Xiang, Xu, Hasan, Hold-Geoffroy, Sunkavalli, and Su]{neutex}
Fanbo Xiang, Zexiang Xu, Milos Hasan, Yannick Hold-Geoffroy, Kalyan Sunkavalli, and Hao Su.
\newblock Neutex: Neural texture mapping for volumetric neural rendering.
\newblock In \emph{Proc. IEEE Conf. Comput. Vis. Pattern Recogn.}, pages 7119--7128, 2021.

\bibitem[Xu et~al.(2022)Xu, Xu, Philip, Bi, Shu, Sunkavalli, and Neumann]{pointnerf}
Qiangeng Xu, Zexiang Xu, Julien Philip, Sai Bi, Zhixin Shu, Kalyan Sunkavalli, and Ulrich Neumann.
\newblock Point-nerf: Point-based neural radiance fields.
\newblock In \emph{Proc. IEEE Conf. Comput. Vis. Pattern Recogn.}, pages 5438--5448, 2022.

\bibitem[Yan et~al.(2020)Yan, Wei, Yi, Ding, Zhang, Chen, Wang, and Tai]{D2HC-RMVSNet}
Jianfeng Yan, Zizhuang Wei, Hongwei Yi, Mingyu Ding, Runze Zhang, Yisong Chen, Guoping Wang, and Yu-Wing Tai.
\newblock Dense hybrid recurrent multi-view stereo net with dynamic consistency checking.
\newblock In \emph{Proc. Eur. Conf. Comput. Vis.}, pages 674--689. Springer, 2020.

\bibitem[Yang et~al.(2020)Yang, Mao, Alvarez, and Liu]{yang2020cost}
Jiayu Yang, Wei Mao, Jose~M Alvarez, and Miaomiao Liu.
\newblock Cost volume pyramid based depth inference for multi-view stereo.
\newblock In \emph{Proc. IEEE Conf. Comput. Vis. Pattern Recogn.}, pages 4877--4886, 2020.

\bibitem[Yao et~al.(2018)Yao, Luo, Li, Fang, and Quan]{mvsnet}
Yao Yao, Zixin Luo, Shiwei Li, Tian Fang, and Long Quan.
\newblock Mvsnet depth inference for unstructured multi-view stereo.
\newblock In \emph{Proc. Eur. Conf. Comput. Vis.}, pages 767--783, 2018.

\bibitem[Yao et~al.(2019)Yao, Luo, Li, Shen, Fang, and Quan]{rmvsnet}
Yao Yao, Zixin Luo, Shiwei Li, Tianwei Shen, Tian Fang, and Long Quan.
\newblock Recurrent mvsnet for high-resolution multi-view stereo depth inference.
\newblock In \emph{Proc. IEEE Conf. Comput. Vis. Pattern Recogn.}, pages 5525--5534, 2019.

\bibitem[Ye et~al.(2023)Ye, Zhao, Liu, Huang, Cao, and Li]{dmvsnet}
Xinyi Ye, Weiyue Zhao, Tianqi Liu, Zihao Huang, Zhiguo Cao, and Xin Li.
\newblock Constraining depth map geometry for multi-view stereo: A dual-depth approach with saddle-shaped depth cells.
\newblock In \emph{Proc. IEEE Int. Conf. Comput. Vis.}, pages 17661--17670, 2023.

\bibitem[Yu et~al.(2021)Yu, Ye, Tancik, and Kanazawa]{pixelnerf}
Alex Yu, Vickie Ye, Matthew Tancik, and Angjoo Kanazawa.
\newblock {pixelNeRF}: Neural radiance fields from one or few images.
\newblock In \emph{Proc. IEEE Conf. Comput. Vis. Pattern Recogn.}, 2021.

\bibitem[Zhang et~al.(2020)Zhang, Yao, Li, Luo, and Fang]{zhang2020visibility}
Jingyang Zhang, Yao Yao, Shiwei Li, Zixin Luo, and Tian Fang.
\newblock Visibility-aware multi-view stereo network.
\newblock \emph{Proc. Br. Mach. Vis. Conf.}, 2020.

\bibitem[Zhang et~al.(2018)Zhang, Isola, Efros, Shechtman, and Wang]{lpips}
Richard Zhang, Phillip Isola, Alexei~A Efros, Eli Shechtman, and Oliver Wang.
\newblock The unreasonable effectiveness of deep features as a perceptual metric.
\newblock In \emph{Proc. IEEE Conf. Comput. Vis. Pattern Recogn.}, pages 586--595, 2018.

\end{thebibliography}
}

\maketitlesupplementary

\section{Appendix} 
In the supplementary material, we present more details that are not included in the main text, including:

\begin{itemize}
    \item More implementation and network details.
    
    \item Additional ablation experiments, including different numbers of views, feature discussions for the CAF module, intermediate view supervision, depth supervisions, and different implementations of ACA networks.

    \item More qualitative results, including novel view, depth maps, fusion weights, and error maps.

    \item Per-scene breakdown results.

    \item Limitations.

\end{itemize}

\setcounter{equation}{0}
\renewcommand{\theequation}{\Roman{equation}}

\setcounter{figure}{0}
\renewcommand{\thefigure}{\Roman{figure}}

\setcounter{table}{0}
\renewcommand{\thetable}{\Roman{table}}

\setcounter{subsection}{0}
\renewcommand{\thesubsection}{\Roman{subsection}}

\begin{figure*}
    \centering
    \includegraphics[width=0.95\textwidth]{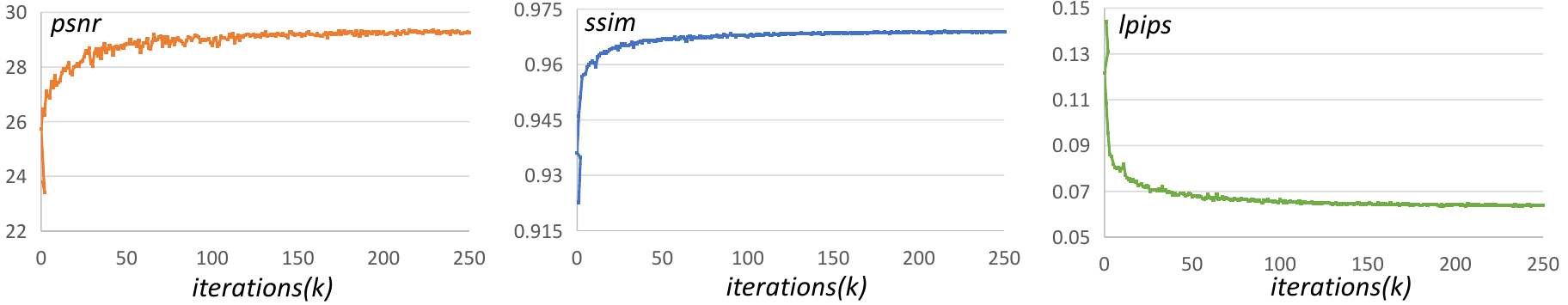}
    \caption{\textbf{The metrics of the DTU test set vary with the number of iterations.}}
    \label{fig:iterations}
\end{figure*}

\subsection{Implementation and Network Details}
\paragraph{Implementation Details.}
Our generalizable model is trained on four RTX $3090$ GPUs using the Adam~\cite{adam} optimizer, with an initial learning rate of $5e-4$. The learning rate is halved every $50$k iterations.
As shown in Fig.~\ref{fig:iterations}, we present the metrics of the DTU test set~\cite{dtu} varying with the number of iterations.
The model tends to converge after approximately $150$k iterations, taking about $25$ hours.
It is worth noting that, with only $8$k iterations, our model can achieve metrics of $27.68/0.961/0.080$, surpassing the metrics of the SOTA method~\cite{enerf}, which is $27.61/0.956/0.091$ (PSNR/SSIM\,\cite{ssim}/LPIPS\,\cite{lpips}).
Following ENeRF~\cite{enerf}, during training, we select $2$, $3$, and $4$ source views as inputs with probabilities of $0.1$, $0.8$, and $0.1$.
To save computational costs, the Consistency-aware Fusion (CAF) is exclusively employed in the fine stage, while in the coarse stage, a blending approach is used to synthesize a low-resolution target view, supervised by the ground-truth target view.
During training, the final target view is generated by fusing two intermediate views as $I_t = W_b I_b + W_r I_r$. However, during evaluation for other datasets~\cite{nerf,llff}, which have a different domain compared with DTU, increasing the weights of $I_b$ can lead to slightly better performance. Therefore, we obtain the final view as $I_t = (I_b + (W_b I_b + W_r I_r))/2$.
Our evaluation setup is consistent with ENeRF~\cite{enerf} and MVSNeRF~\cite{mvsnerf}.
The results of the DTU test set are evaluated using segmentation masks. The segmentation mask is defined based on the availability of ground-truth depth at each pixel.
Since the marginal region of images is typically invisible to input images on the Real Forward-facing dataset~\cite{llff}, we evaluate the $80\%$ area in the center of the images.
Incidentally, all the inference time presented in Fig.~1 of the main text is measured at an input image resolution of $512\times640$. The number of our model's parameters is $3.15$M.

\begin{table}
\centering
\renewcommand\arraystretch {1}
\resizebox{0.7\linewidth}{!}{
\begin{tabular}{@{}c|c|c@{}}
\toprule
Input & Operation & Output \\
\midrule
(C,D,H,W) & permute & (D,C,H,W) \\
(D,C,H,W) & cgr(C,C$_h$) & (D,C$_h$,H,W) \\
(D,C$_h$,H,W) & cgr(C$_h$,C$_h$) & (D,C$_h$,H,W) \\
(D,C$_h$,H,W) & conv2d(C$_h$,1) & (D,1,H,W) \\
(D,1,H,W) & sigmoid & (D,1,H,W) \\ 
(D,1,H,W) & permute & (1,D,H,W) \\
\bottomrule
\end{tabular}}
\caption{\textbf{The network architecture of Adaptive Cost Aggregation (ACA).} The cgr represents a block composed of conv2d, groupnorm, and relu. In our implementation, $C=16$ and $C_h=4$.}
\label{tab:aca_network}
\end{table}

\paragraph{Network Details.}
Here, we will introduce the network details of the pooling network (Sec.~4.2) and ACA (Sec.~4.1) mentioned in the main text.

Pooling network. ~\cite{ibrnet,enerf} apply a pooling network $\rho$ to aggregate multi-view features to obtain the descriptor via $f_p = \rho(\{f_s^i\}_{i=1}^N)$.
The implementation details are as follows: initially, the mean $\mu$ and variance $v$ of $\{f_s^i\}_{i=1}^N$ are computed. Subsequently, $\mu$ and $v$ are concatenated with each $f_s^i$ and an MLP is applied to generate a weight. The $f_p$ is blended via a soft-argmax operator using obtained weights and multi-view features $(\{f_s^i\}_{i=1}^N)$.

ACA. Per the Eq.~(4) in the main text, $\alpha(.)$ represents the adaptive weight for each view, and the network architecture that learns these weights is shown in Table~\ref{tab:aca_network}.

\paragraph{Inference Speed.} 
For an image with $512\times640$ resolution, the inference time of our method is $143$ms. We decompose the inference time in Table~\ref{Table:time} and results demonstrate that the inference time of our modules is only $71$ms, with the remaining $72$ms spent saving results.

\begin{table}
\centering
\resizebox{0.65\linewidth}{!}{
\begin{tabular}{@{}cccc@{}}
\toprule
\multicolumn{2}{c}{Modules} & coarse stage & fine stage \\
\midrule
\multicolumn{2}{c}{feature extractor} &  \multicolumn{2}{c}{1.32} \\
\midrule
\multicolumn{2}{c}{build rays} & 9.11 & 15.94 \\
\midrule
\multirow{2}{*}{geometry} & cost volume & 7.48 & 8.24 \\
& regularization & 1.91 & 1.97 \\
\midrule
\multirow{3}{*}{descriptor} & pooling $\rho$ & 0.57 & 0.50 \\
& SVA$_{sm}$ & 1.74 & 1.37 \\
& SVA$_{d}$ & 0.73 & 1.01 \\
\midrule
\multicolumn{2}{c}{view decoding} &  1.62 & 17.80 \\
\midrule
\multicolumn{2}{c}{save in dictionary} & 13.27 & 58.31 \\
\bottomrule
\end{tabular}
}
\caption{\textbf{Time overhead for each module (in milliseconds).} The term ``save in dictionary'' refers to storing tensor results in dictionary form for subsequent evaluation of various metrics.}
\label{Table:time}
\end{table}

\begin{table*}
\centering
\renewcommand\arraystretch {1}
\resizebox{0.85\linewidth}{!}{
\begin{tabular}{@{}c|cccccc@{}}
\toprule
Views & PSNR$\uparrow$ & SSIM$\uparrow$ & LPIPS$\downarrow$ & Abs err$\downarrow$ & Acc(2)$\uparrow$ & Acc(10)$\uparrow$ \\
\midrule
2 & 26.98 / 25.48 & 0.955 / 0.942 & 0.081 / 0.107 & 3.86 / 5.53 & 0.835 / 0.756 & 0.942 / 0.107 \\
3 & 29.36 / 27.61 & 0.969 / 0.957 & 0.064 / 0.089 & 2.83 / 4.60 & 0.879 / 0.792 & 0.961 / 0.917 \\
4 & 29.77 / 27.73 & 0.971 / 0.959  & 0.062 / 0.089 & 2.73 / 4.26 & 0.880 / 0.804 & 0.961 / 0.929 \\
5 & 29.91 / 27.54 & 0.971 / 0.958 & 0.062 / 0.091 & 2.69 / 4.29 & 0.882 / 0.800 & 0.961 / 0.928 \\
\bottomrule
\end{tabular}}
\caption{\textbf{The performance of our method and ENeRF with different numbers of input views on the DTU test set.} Each item represents (Ours/ENeRF's). ``Abs err'' denotes the average absolute error and ``Acc(X)'' means the
percentage of pixels with an error less than X mm.}
\label{tab:ab_view_dtu}
\end{table*}

\begin{table}
\centering
\setlength{\tabcolsep}{5pt}
\resizebox{0.99\linewidth}{!}{
\begin{tabular}{@{}ccccccc@{}}
\toprule
\multirow{2}{*}{Views} & \multicolumn{3}{c}{Real Forward-facing~\cite{llff}} & \multicolumn{3}{c}{NeRF Synthetic~\cite{nerf}}\\ 
\cmidrule(lr){2-4}\cmidrule(lr){5-7}
 & PSNR $\uparrow$ & SSIM $\uparrow$ & LPIPS $\downarrow$ & PSNR $\uparrow$ & SSIM $\uparrow$ & LPIPS $\downarrow$ \\
\midrule
2 & 23.39 & 0.839 & 0.176 & 25.30 & 0.939 & 0.082 \\
3 & 24.28 & 0.863 & 0.162 & 26.99 & 0.952 & 0.070 \\
4 & 24.91 & 0.876 & 0.157 & 27.31 & 0.953 & 0.069 \\
\bottomrule
\end{tabular}}
\caption{\textbf{The performance of our method with varying numbers of input views on the Real Forward-facing and NeRF Synthetic datasets.}}
\label{tab:ab_view_llff_nerf}
\end{table}

\subsection{Additional ablation experiments}
\paragraph{Numbers of Views.}
As shown in Table~\ref{tab:ab_view_dtu}, we evaluate the performance of our trained generalization model and ENeRF~\cite{enerf} with different numbers of input views on the DTU test set~\cite{dtu}.
With an increase in the number of input views, the performance improves as the model can leverage 
more multi-view information.
In terms of both overall performance and the magnitude of performance improvement, our method outperforms ENeRF, indicating its superior capability in leveraging multi-view information for reconstructing scene geometry and rendering novel views.
Additionally, we also present the performance of our method on the Real Forward-facing~\cite{llff} and NeRF Synthetic~\cite{nerf} datasets under different numbers of input views, as shown in Table~\ref{tab:ab_view_llff_nerf}. 
The results demonstrate the same trend, indicating the capability of our model in leveraging multi-view information is generalizable.

\paragraph{Features for Intermediate Views.}
In Sec.~4.3 of the main text, $f_b$ and $f_r$ are utilized as the feature representations for the two intermediate views $I_b$ and $I_r$, respectively. Subsequently, their consistency with source views is individually computed to learn fusion weights.
The choice of using $f_b$ as the feature for $I_b$ is based on their similar volume rendering generation manners, while the selection of $f_r$ as the feature for $I_r$ is driven by their direct projection relationship.
Here, we will discuss different selection strategies for the features of intermediate views.
An alternative approach for the features of $I_b$ is to blend features from source views.
Similar to Eq.~(2) in the main text, the calculation of the features $f_b$ for $I_b$ is as follows:
\begin{equation}
\label{eq:blend_feature}
f_b = \sum_{i=1}^{N} \frac{exp(w_i) f_s^i}{\sum_{j=1}^{N} exp(w_j)} \,,
\end{equation}
where $w_i = \text{MLP} (x,d,f_p, f_s^i)$, $f_s^i$ is the feature of the source image $I_s^i$. $x$ and $d$ represent the coordinate and view direction, respectively. $f_p$ is the descriptor for 3D point.
Another more intuitive alternative is to use a feature extractor to extract features for both intermediate views, as:
\begin{equation}
\label{eq:extract_feature}
F_{[b,r]} = \phi_e I_{[b,r]} \,,
\end{equation}
where $\phi_e$ represents a feature extractor, instantiated as a 2D U-Net. $f_b$ and $f_r$ are the pixel-wise features of $F_b$ and $F_r$, respectively.
As shown in Table~\ref{tab:intermediate_feature}, the strategy employed in the main text is slightly superior to the other two alternative strategies.
For the first alternative~\cref{eq:blend_feature}, $f_b$ is obtained by blending features from source views. $f_b$ lacks 3D context awareness, leading to some information loss in the subsequently accumulated pixel features.
For the second alternative~\cref{eq:extract_feature}, $f_b$ and $f_r$ are extracted from scratch at the RGB level.
This practice is disadvantageous for the subsequent learning of 3D consistency weights, due to the lack of utilization of 3D information.
Additionally, the introduction of a feature extractor also increases the burden on the model.
However, the strategy in the main text maximally utilizes the obtained 3D-aware descriptors, while also having the smallest computational cost compared to the other two alternative approaches.

\begin{table}
\centering
\renewcommand\arraystretch {1}
\resizebox{0.55\linewidth}{!}{
\begin{tabular}{@{}c|ccc@{}}
\toprule
 & PSNR$\uparrow$ & SSIM$\uparrow$ & LPIPS$\downarrow$  \\
\midrule
No.1 & 29.36 & 0.969 & 0.064 \\
No.2 & 29.24 & 0.969 & 0.065 \\
No.3 & 29.08 & 0.968 & 0.066 \\
\bottomrule
\end{tabular}}
\caption{\textbf{Different strategies for the features of intermediate views.} No.1 represents the strategy in the main text. No.2 represents the strategy using \cref{eq:blend_feature}. No.3 represents the strategy using \cref{eq:extract_feature}.}
\label{tab:intermediate_feature}
\end{table}

\begin{table}
\centering
\renewcommand\arraystretch {1}
\resizebox{0.99\linewidth}{!}{
\begin{tabular}{@{}c|cccccc@{}}
\toprule
 Depth & PSNR$\uparrow$ & SSIM$\uparrow$ & LPIPS$\downarrow$ & Abs err$\downarrow$ & Acc(2)$\uparrow$ & Acc(10)$\uparrow$ \\
\midrule
Self-supervision & 29.21 & 0.968 & 0.064 & 3.21 & 0.873 & 0.957 \\
Supervision & 29.31 & 0.969 & 0.064 & 2.95 & 0.875 & 0.957 \\
None & 29.36 & 0.969 & 0.064 & 2.83 & 0.879 & 0.961 \\
\bottomrule
\end{tabular}}
\caption{\textbf{The comparison of different depth supervision signals.} The self-supervision represents using unsupervised depth loss and the supervision represents using ground-truth depth for supervision. The term ``None'' refers to training without any depth supervision signals.}
\label{tab:ab_depth_supervision}
\end{table}

\paragraph{Intermediate View Supervision.}
In the main text, we only supervise the images fused through CAF.
However, simultaneously supervising intermediate results is also a common practice, whose final result is $29.15/0.968/0.065$.
This result is slightly inferior to supervising only the fused view ($29.36/0.969/0.064$).
Because each of the two intermediate views has its own advantages, supervising only the fused view allows the network to focus on the fusion process, leveraging the strengths of both.
However, simultaneously supervising the intermediate views burdens the network, diminishing its attention to the fusion process. In theory, if both intermediate views are entirely correct, the final fused view will be accurate regardless of the fusion process. The network prioritizes predicting two accurate intermediate views, which is a more challenging task.

\paragraph{Depth Supervision.}
A critical factor in the model's synthesis of high-quality views is its perception of the scene geometry.
MVS-based generalizable NeRF methods~\cite{mvsnerf,enerf,neuray}, including our method, aim to improve the quality of view synthesis by enhancing the geometry prediction.
By only supervising RGB images, excellent geometry predictions can be achieved (Sec.~5.4 in the main text).
Here, we will discuss the impact of incorporating depth supervision signals on the model.
We introduce supervision signals for depth in two ways: one through self-supervised and the other through supervision using ground-truth depth.

Following~\cite{khot2019learning,RC-MVSNet}, the unsupervised depth loss is:
\begin{equation}
\label{eq:unsupervised depth loss}
L_d = \beta_1 L_{PC} + \beta_2 L_{SSIM} + \beta_3 L_{Smooth} \,,
\end{equation}
where $L_{PC}$ represents the photometric consistency loss. $L_{SSIM}$ and $L_{Smooth}$ are the structured similarity loss and depth smoothness loss, respectively. 
$\beta_1$, $\beta_2$, and $\beta_3$ are set to $12$, $6$, and $0.18$ in our implementation, respectively. Refer to~\cite{khot2019learning,RC-MVSNet} for more details.
$L_d$ is used to supervise the final depth, \ie, $d_f$ (Sec.~4.3 in the main text).
Since the DTU dataset provides ground-truth depth, another approach is to utilize the ground-truth depth for supervision.
The depth loss is as follows:
\begin{equation}
\label{eq:supervised depth loss}
L_d = \xi(d_f,d_{gt})
\end{equation}
where $d_f$ and $d_{gt}$ represent the final predicted and ground-truth depth, respectively. $\xi$ denotes a loss function. Following~\cite{gu2020cascade}, $\xi$ is instantiated as the Smooth L1 loss~\cite{FastR-CNN}.
The quantitative results are presented in the Table~\ref{tab:ab_depth_supervision}.
The performance of the three strategies in the table is comparable, indicating that supervising only the RGB images is sufficient, and there is no need for additional introduction of depth supervision signals.

\paragraph{More Comprehensive Depth Analysis.}
As shown in Fig.~3 in the main text, our pipeline first infers the geometry from the cost volume, re-samples 3D points around objects' surfaces, and finally encodes 3D descriptors for rendering.
We can obtain two depths: one inferred from the cost volume and the other obtained through volume rendering, which is the final depth.
Here, we report the depth obtained from the cost volume and the final depth as shown in Table \ref{Tab:mvs_depth}.
Compared to the baseline, our method performs better on both depth metrics.
Thanks to Adaptive Cost Aggregation (ACA), the depth obtained from the cost volume has been significantly improved.
Based on this, as the Spatial-View Aggregator (SVA) encodes 3D-aware descriptors, the final depth has also been further improved.
In addition, the well-designed decoding approach, \ie, CAF, greatly facilitates the depth prediction of the model (Sec.~5.5 in the main text).

\begin{figure}
    \centering
    \includegraphics[width=0.49\textwidth]{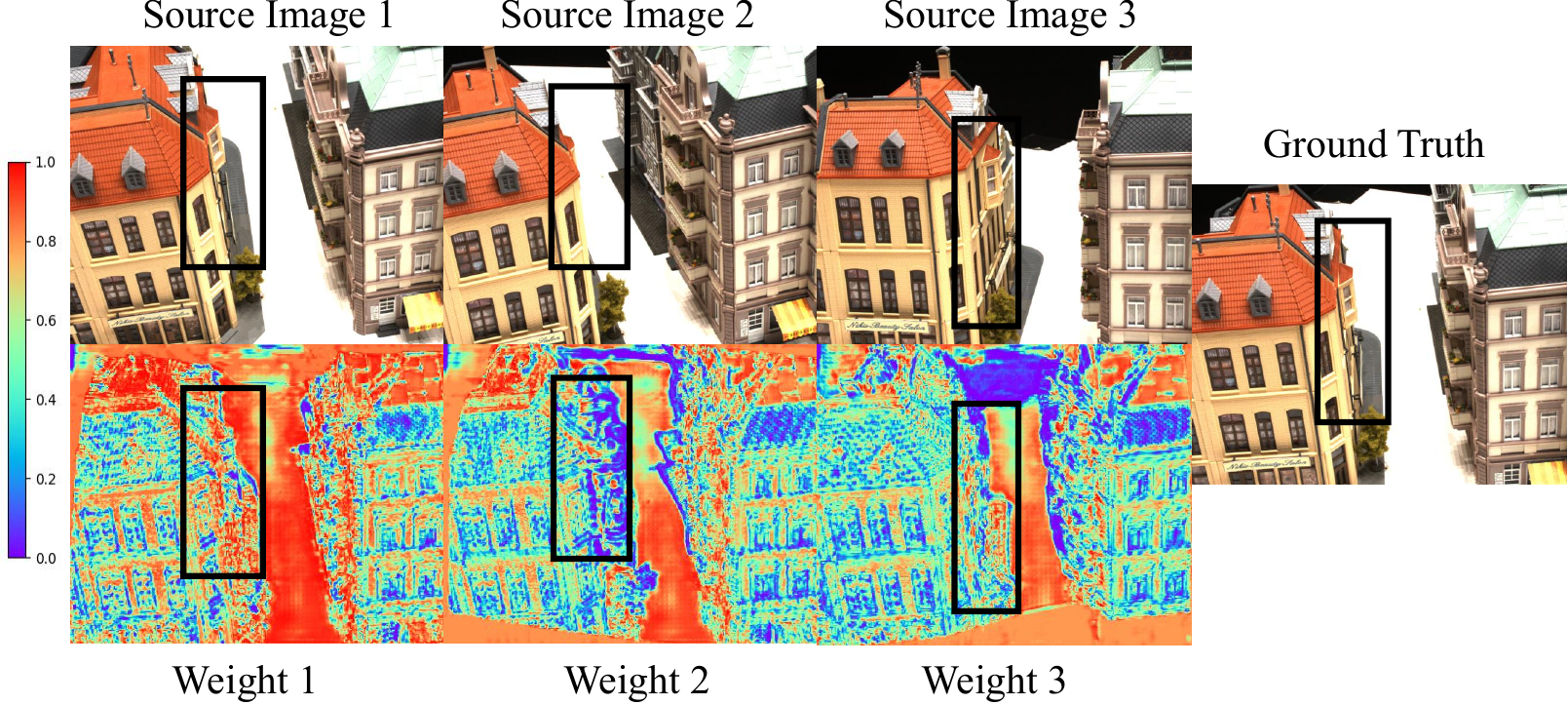}
    \caption{\textbf{Visualization of ACA.}}
    \label{fig:vis_aca}
\end{figure}

\paragraph{Visualization of ACA.}
Previous approaches using variance struggle to encode efficient cost information for challenging areas, such as the occluded areas marked in the black box in Fig.~\ref{fig:vis_aca}.
Our proposed ACA module learns an adaptive weight for different views to encode accurate geometric information.
As illustrated in Fig.~\ref{fig:vis_aca}, the weights learned by ACA amplify the contribution of consistent pixel pairs, such as the visible areas in source image $1$ and $3$, while suppressing inconsistent ones, as shown in the occluded areas in the source image $2$.

\paragraph{Different ACA networks.}
The primary challenge of applying ACA to the NVS task is the unavailability of the target view, which we addressed by adopting a coarse-to-fine framework.
In the main text, the weight learning network utilized in ACA is illustrated in Table~\ref{tab:aca_network}, following the MVS method, \ie, AA-RMVSNet~\cite{aa-rmvsnet}.
Moreover, other networks can also be embedded into our coarse-to-fine framework to learn inter-view weights.
Here, we adopt another MVS method, \ie, MVSTER~\cite{mvster}, to learn adaptive weights.
The result on the DTU test set is $29.31/0.969/0.064$ (PSNR/SSIM/LPIPS), which is comparable with the result obtained using~\cite{aa-rmvsnet}.
In summary, our main contribution is to propose an approach for applying ACA to the NVS task, without specifying a particular network for learning weights.

\begin{figure}
    \centering
    \includegraphics[width=0.49\textwidth]{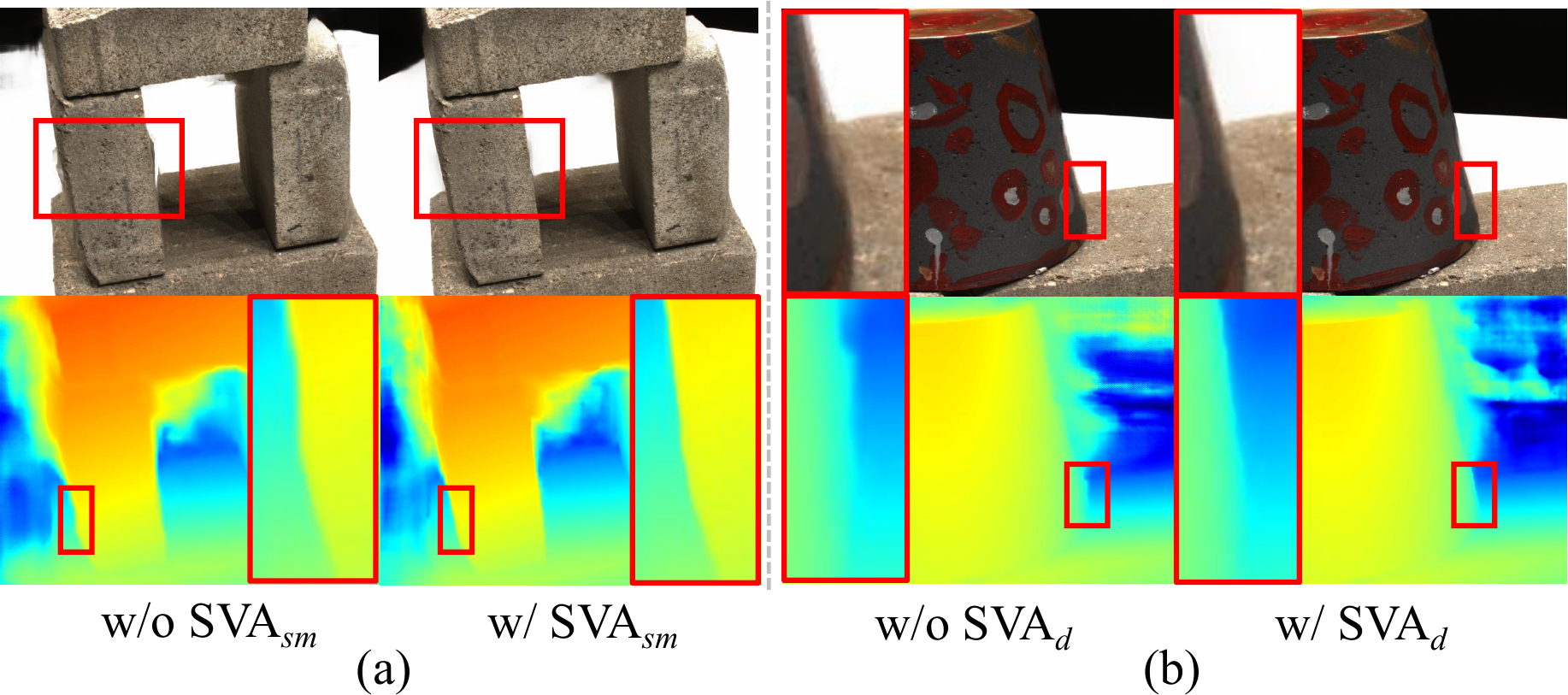}
    \caption{Visualization of SVA. SVA$_{sm}$ and SVA$_{d}$ represent $\phi_{sm}$ and $\phi_{d}$ of SVA, respectively.}
    \label{fig:vis_sva}
\end{figure}

\paragraph{Analysis of SVA.}
Previous approaches directly uses a pooling network to aggregate multi-view 2D features for encoding 3D descriptors, which are not spatially context-aware, leading to discontinuities in the decoded depth map and rendered view (see Fig.~\ref{fig:vis_sva} (a)). 
To address this issue, convolutional networks can be used to impose spatial constraints on adjacent descriptors. However, due to the smooth nature of convolution, some high-frequency details may be lost. Since detailed information comes from the multi-view features, we employ a divide-and-conquer approach to aggregate descriptors.
Firstly, we employ a 3D U-Net to aggregate spatial context and obtain smooth descriptors. Despite resolving the issue of discontinuities, an unsharpened object edge occurs (Fig.~\ref{fig:vis_sva} (b)).
Secondly, we propose using smoothed features as queries, with multi-view features serving as keys and values. Applying the attention mechanism allows us to gather high-frequency details adaptively. This practice results in continuities and sharp boundaries in both rendered views and depth maps.

\begin{table}
\centering
\setlength{\tabcolsep}{1pt}
\renewcommand\arraystretch {1}
\scalebox{1}{
\resizebox{\linewidth}{!}{
\begin{tabular}{@{}lcccccc@{}}
\toprule
\multirow{2}{*}{Method} & \multicolumn{3}{c}{Reference view} & \multicolumn{3}{c}{Novel view} \\ 
\cmidrule(lr){2-4}\cmidrule(lr){5-7}
& Abs err $\downarrow$ & Acc(2)$\uparrow$ & Acc(10)$\uparrow$ & Abs err $\downarrow$ & Acc(2)$\uparrow$ & Acc(10)$\uparrow$ \\
\midrule
Baseline-mvs & 3.71 & 0.815 & 0.942 & 4.49 & 0.778 & 0.928 \\
Baseline-final & 3.58 & 0.842 & 0.944 & 4.32 & 0.800 & 0.928 \\ 
Ours-mvs & 2.79 & 0.836 & 0.965 & 3.15 & 0.816 & 0.958 \\
Ours-final & 2.47 & 0.900 & 0.971 & 2.83 & 0.879 & 0.961 \\
\bottomrule
\end{tabular}}
}
\caption{\textbf{More comprehensive depth metrics.}``-mvs'' represents the depth obtained from the cost volume and ``-final'' represents the final depth obtained through volume rendering.}
\label{Tab:mvs_depth}
\end{table}

\begin{table}
\centering
\renewcommand\arraystretch {1}
\resizebox{0.65\linewidth}{!}{
\begin{tabular}{@{}c|ccc@{}}
\toprule
Approach & PSNR$\uparrow$ & SSIM$\uparrow$ & LPIPS$\downarrow$  \\
\midrule
Regression & 27.32 & 0.945 & 0.119 \\
Blending & 28.40 & 0.962 & 0.091 \\
Overall & 29.36 & 0.969 & 0.064 \\
\bottomrule
\end{tabular}}
\caption{\textbf{Quantitative results for intermediate results.} Overall represents the fused views.}
\label{tab:intermediate_metrics}
\end{table}

\subsection{More Qualitative Results}
\begin{figure*}
    \centering
    \includegraphics[width=0.8\textwidth]{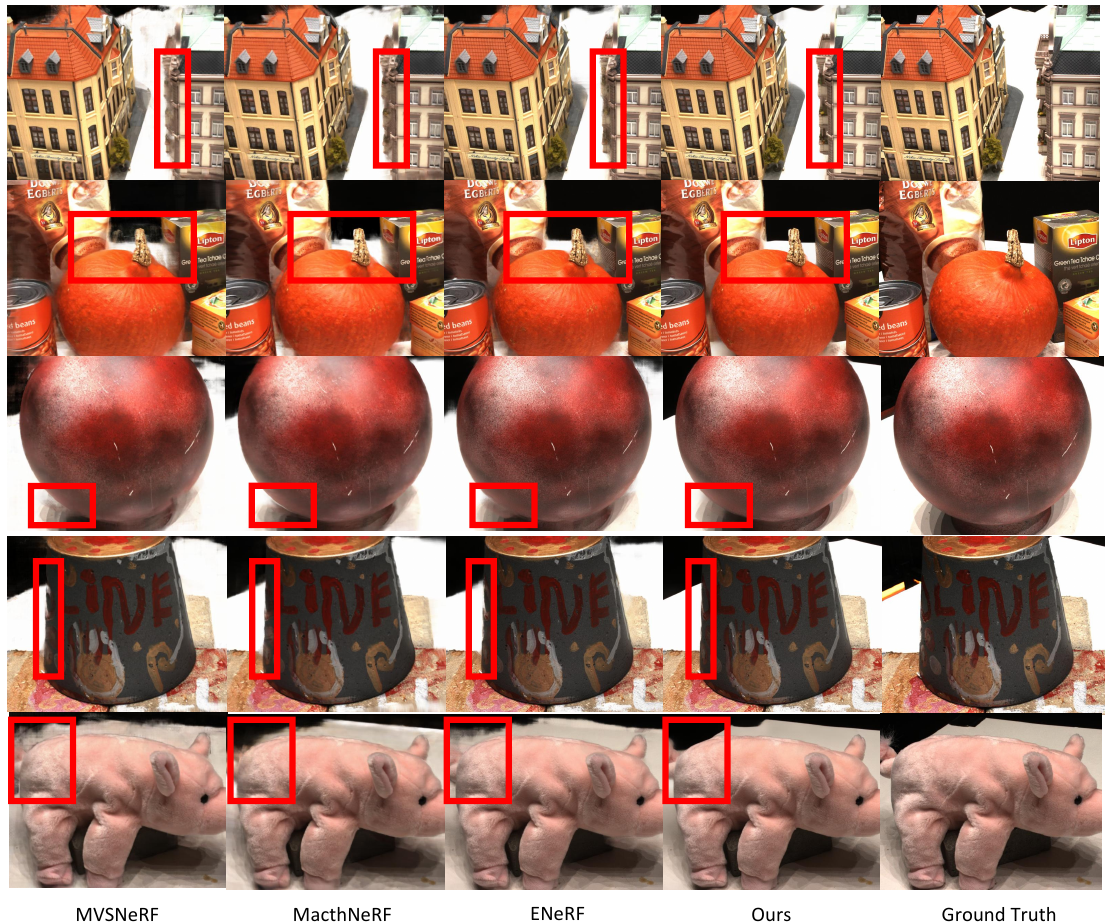}
    \caption{\textbf{Qualitative comparison of rendering quality with state-of-the-art methods~\cite{mvsnerf,matchnerf,enerf} on the DTU dataset under generalization and three views settings.}}
    \label{fig:gen_vis_dtu}
\end{figure*}

\begin{figure*}
    \centering
    \includegraphics[width=0.95\textwidth]{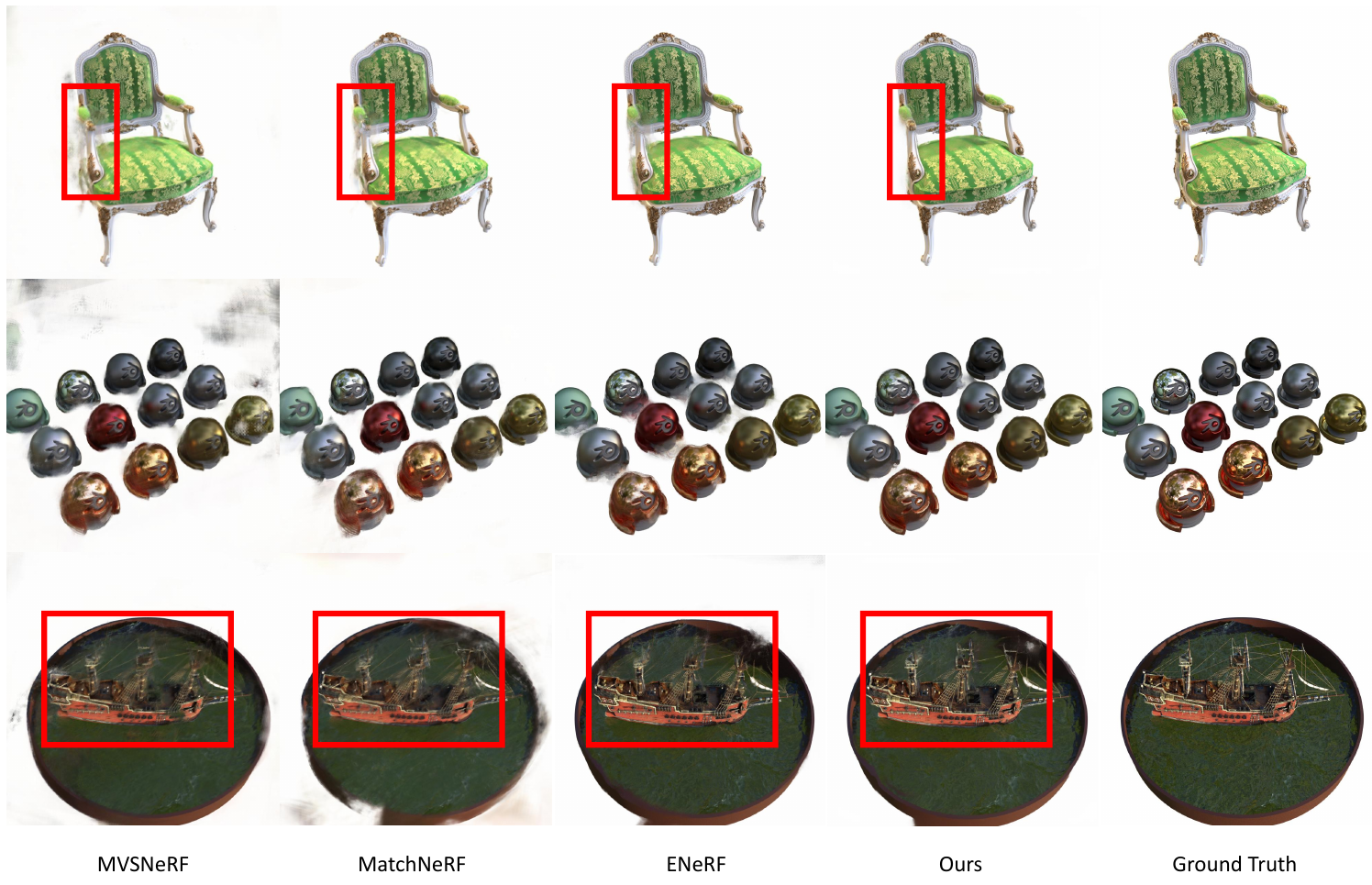}
    \caption{\textbf{Qualitative comparison of rendering quality with state-of-the-art methods~\cite{mvsnerf,matchnerf,enerf} on the NeRF Synthetic dataset under generalization and three views settings.}}
    \label{fig:gen_vis_nerf}
\end{figure*}

\begin{figure*}
    \centering
    \includegraphics[width=0.95\textwidth]{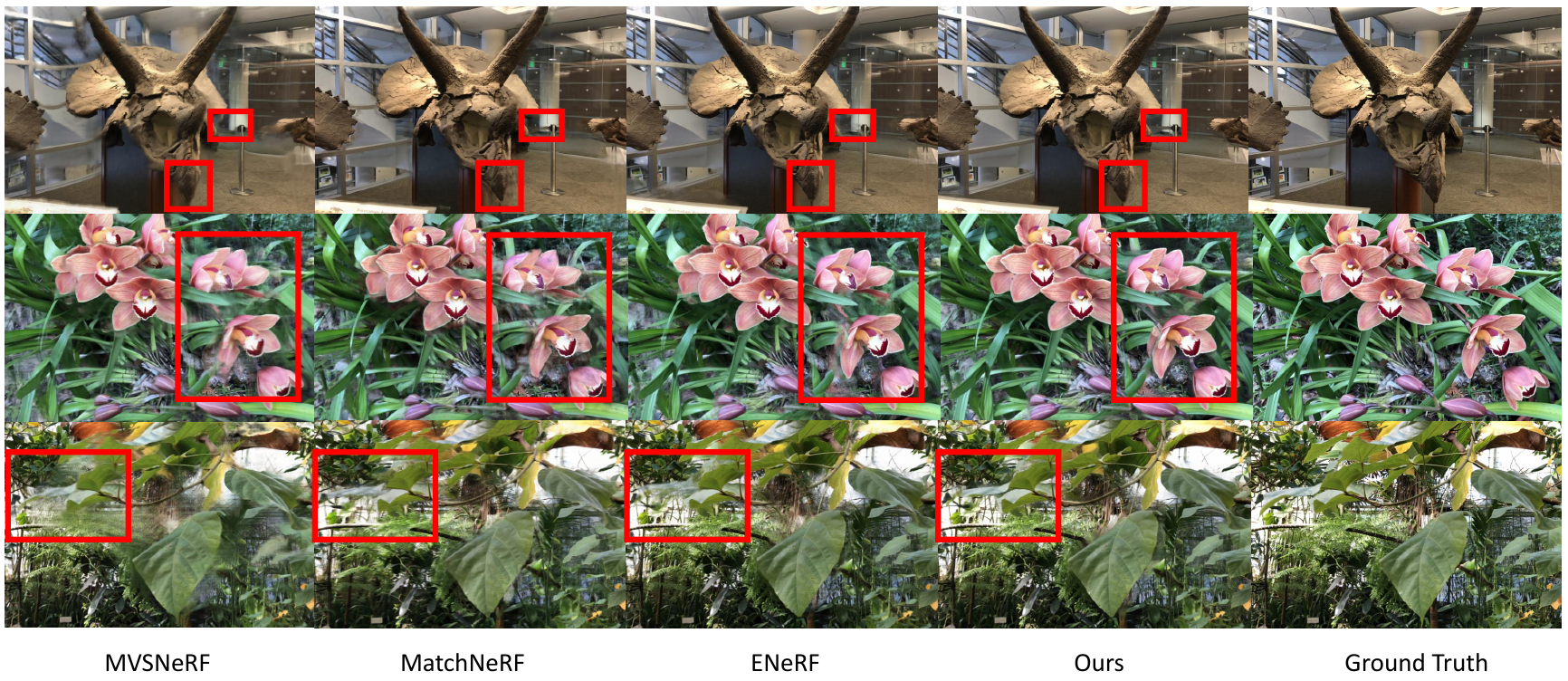}
    \caption{\textbf{Qualitative comparison of rendering quality with state-of-the-art methods~\cite{mvsnerf,matchnerf,enerf} on the Real Forward-facing dataset under generalization and three views settings.}}
    \label{fig:gen_vis_llff}
\end{figure*}

\paragraph{Qualitative Results under the Generalization Setting.}
As shown in Fig.~\ref{fig:gen_vis_dtu},~\ref{fig:gen_vis_nerf}, and \ref{fig:gen_vis_llff}, we present the qualitative comparison of rendering quality on the DTU~\cite{dtu}, NeRF Synthetic~\cite{nerf}, and Real Forward-facing~\cite{llff} datasets, respectively.
Our method can synthesize views with higher fidelity, especially in challenging areas. For example, in occluded regions and geometrically complex scenes, our method can reconstruct more details while exhibiting fewer artifacts at objects' edges and in reflective areas.

\begin{figure*}
    \centering
    \includegraphics[width=0.95\textwidth]{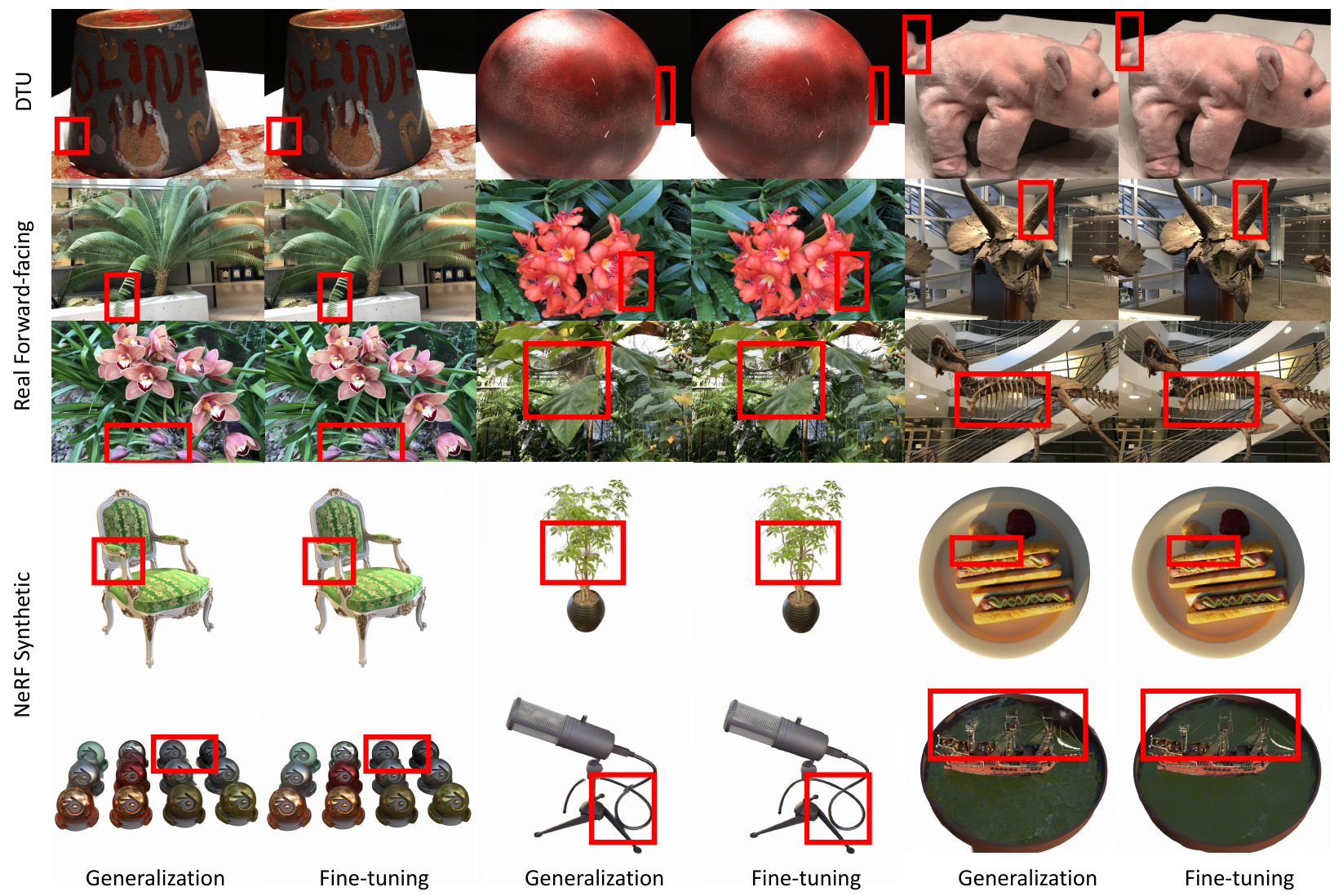}
    \caption{\textbf{Qualitative comparison of results before and after fine-tuning on the DTU~\cite{dtu}, Real Forward-facing~\cite{llff}, and NeRF Synthetic~\cite{nerf} datasets.}}
    \label{fig:gen_finetuning}
\end{figure*}

\paragraph{Qualitative Results under the Per-scene Optimization Setting.}
Benefiting from the strong initialization of our generalizable model, excellent performance can be achieved within just a short fine-tuning period, such as 15 minutes.
As shown in Fig.~\ref{fig:gen_finetuning}, we present the results after fine-tuning. 
After per-scene optimization, the model demonstrates enhanced capabilities in handling scene details, resulting in views with higher fidelity.

\begin{figure*}
    \centering
    \includegraphics[width=0.90\textwidth]{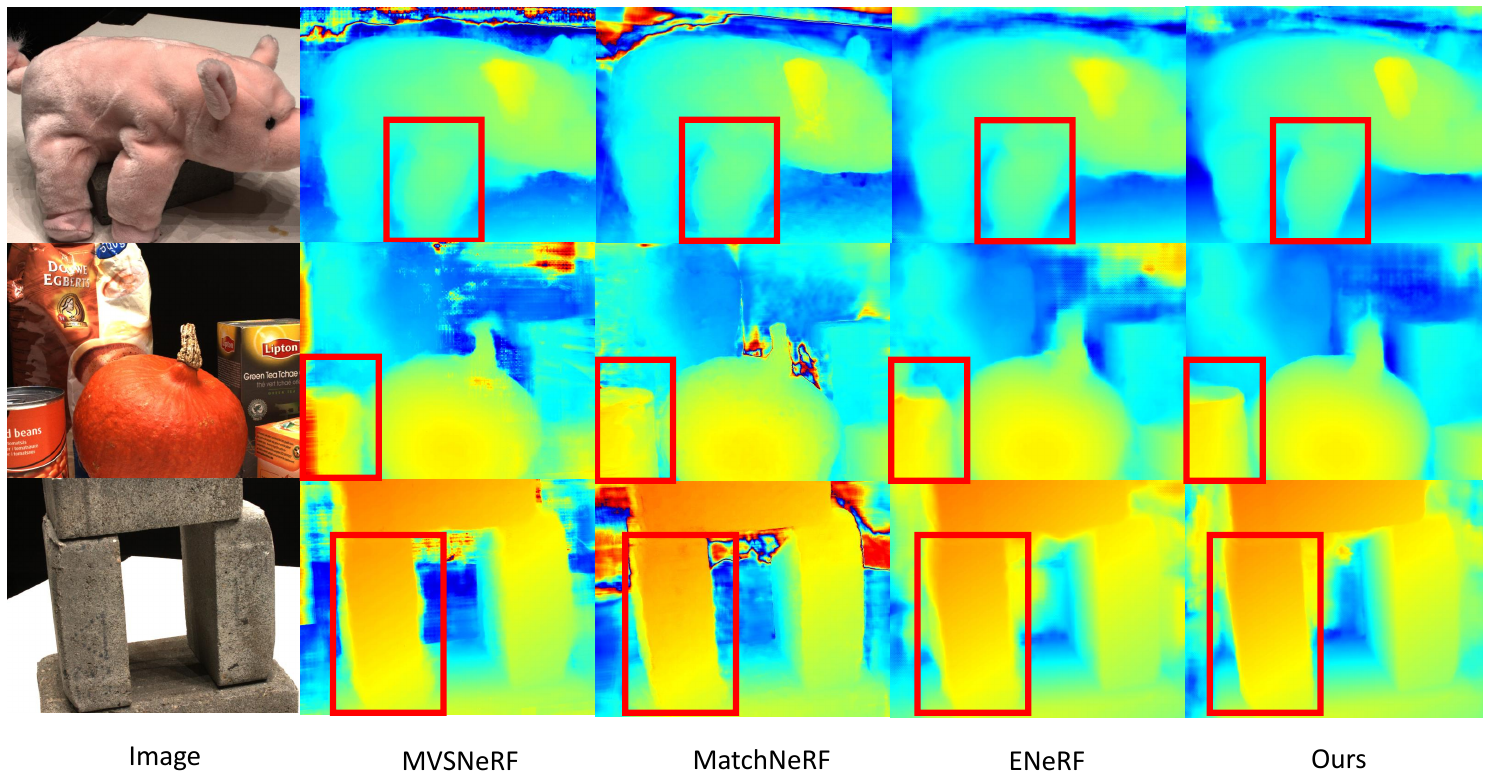}
    \caption{\textbf{Qualitative comparison of depth maps with state-of-the-art methods~\cite{mvsnerf,matchnerf,enerf} on the DTU dataset.}}
    \label{fig:depth_vis_dtu}
\end{figure*}

\begin{figure*}
    \centering
    \includegraphics[width=0.85\textwidth]{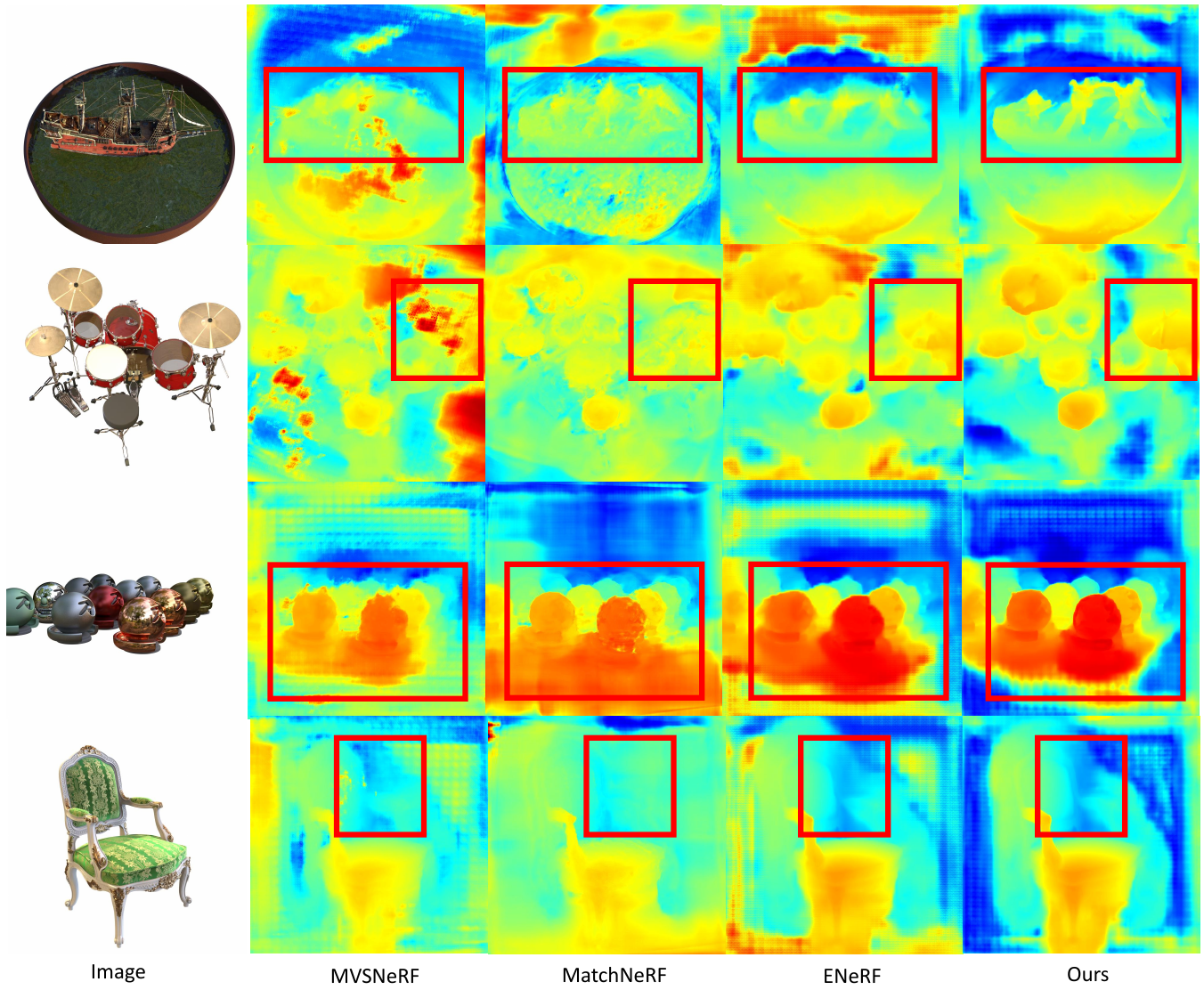}
    \caption{\textbf{Qualitative comparison of depth maps with state-of-the-art methods~\cite{mvsnerf,matchnerf,enerf} on the NeRF Synthetic dataset.}}
    \label{fig:depth_vis_nerf}
\end{figure*}

\begin{figure*}
    \centering
    \includegraphics[width=0.85\textwidth]{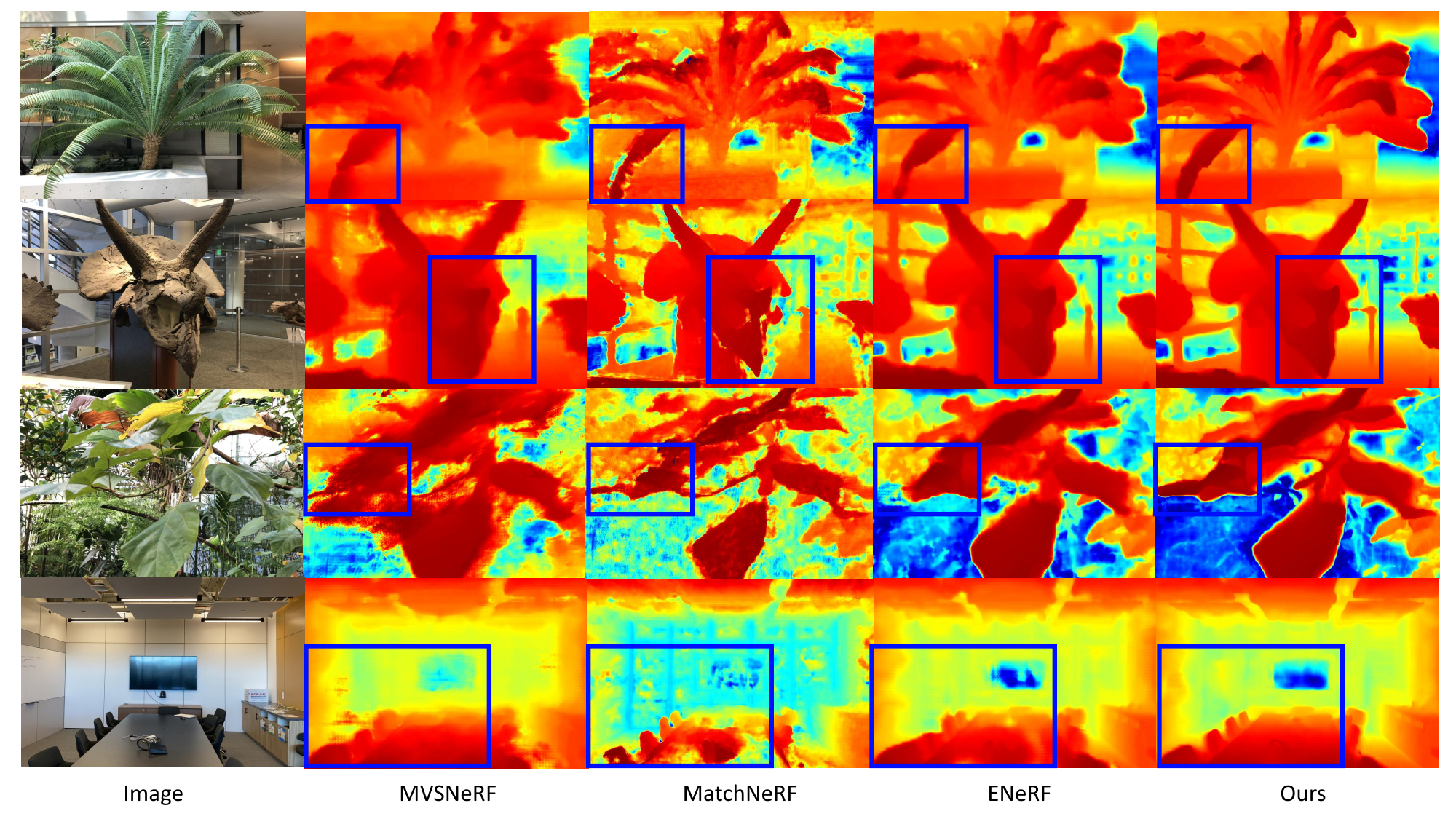}
    \caption{\textbf{Qualitative comparison of depth maps with state-of-the-art methods~\cite{mvsnerf,matchnerf,enerf} on the Real Forward-facing dataset.}}
    \label{fig:depth_vis_llff}
\end{figure*}

\paragraph{Qualitative Comparison of Depth Maps.}
As shown in Figs.~\ref{fig:depth_vis_dtu},~\ref{fig:depth_vis_nerf}, and \ref{fig:depth_vis_llff}, we present the qualitative comparison of depth maps on the DTU~\cite{dtu}, NeRF Synthetic~\cite{nerf}, and Real Forward-facing~\cite{llff} datasets, respectively.
The depth maps generated by our method can maintain sharper object edges and preserve more details of scenes, which verifies the strong geometry reasoning capability of our method. 

\begin{figure*}
    \centering
    \includegraphics[width=0.95\textwidth]{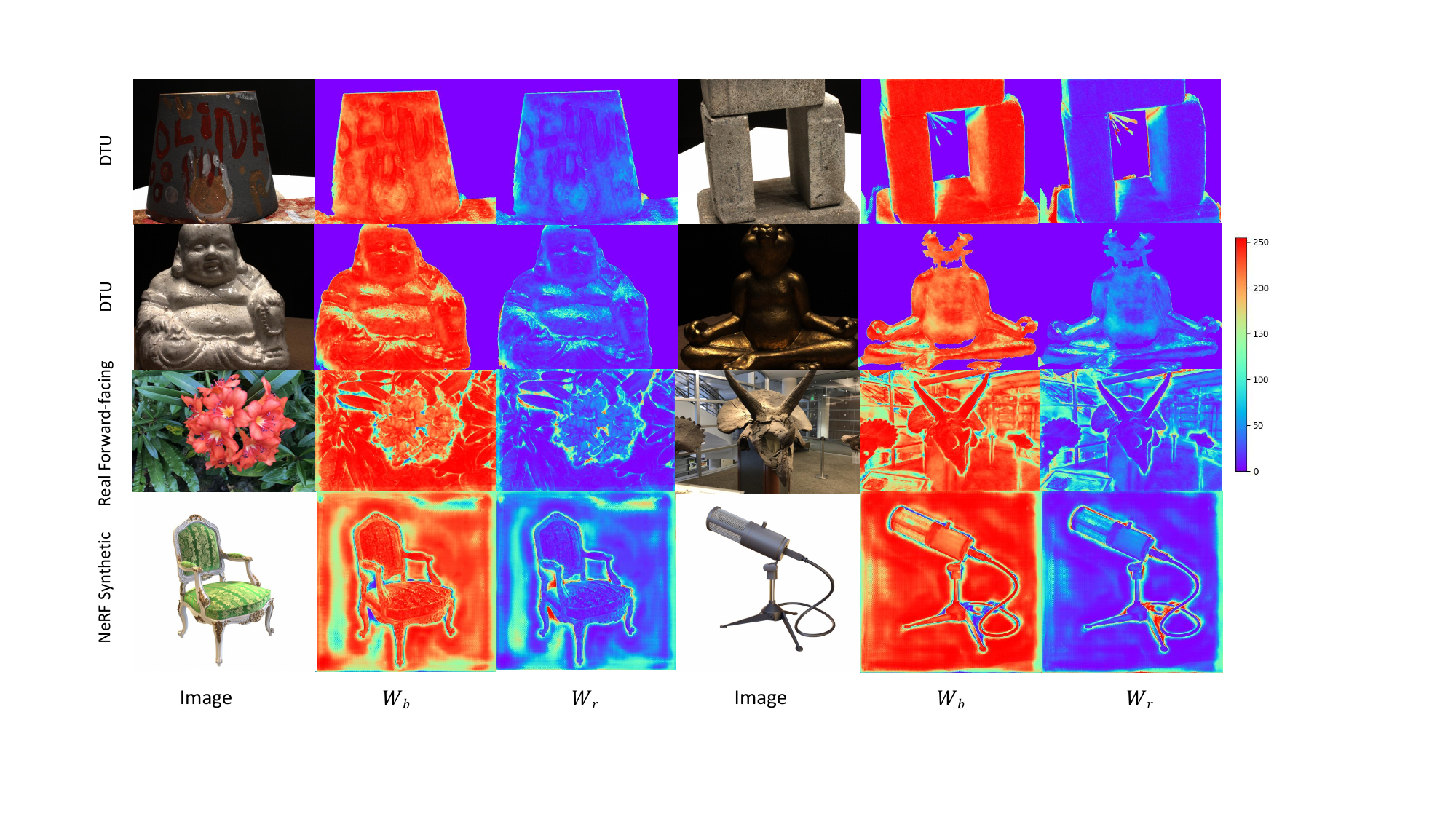}
    \caption{\textbf{Visualization of Fusion Weights.} $W_b$ and $W_r$ represent the weight maps of the blending approach and regression approach, respectively.}
    \label{fig:supp_vis_weights}
\end{figure*}

\paragraph{Fusion Weight Visualization.}
As shown in Fig.~\ref{fig:supp_vis_weights}, we present the fusion weights of the Consistency-aware Fusion (CAF) module.
The blending approach generally demonstrates higher confidence in most areas, while the regression approach shows higher confidence in challenging regions such as object boundaries and reflections.

\begin{figure*}
    \centering
    \includegraphics[width=0.95\textwidth]{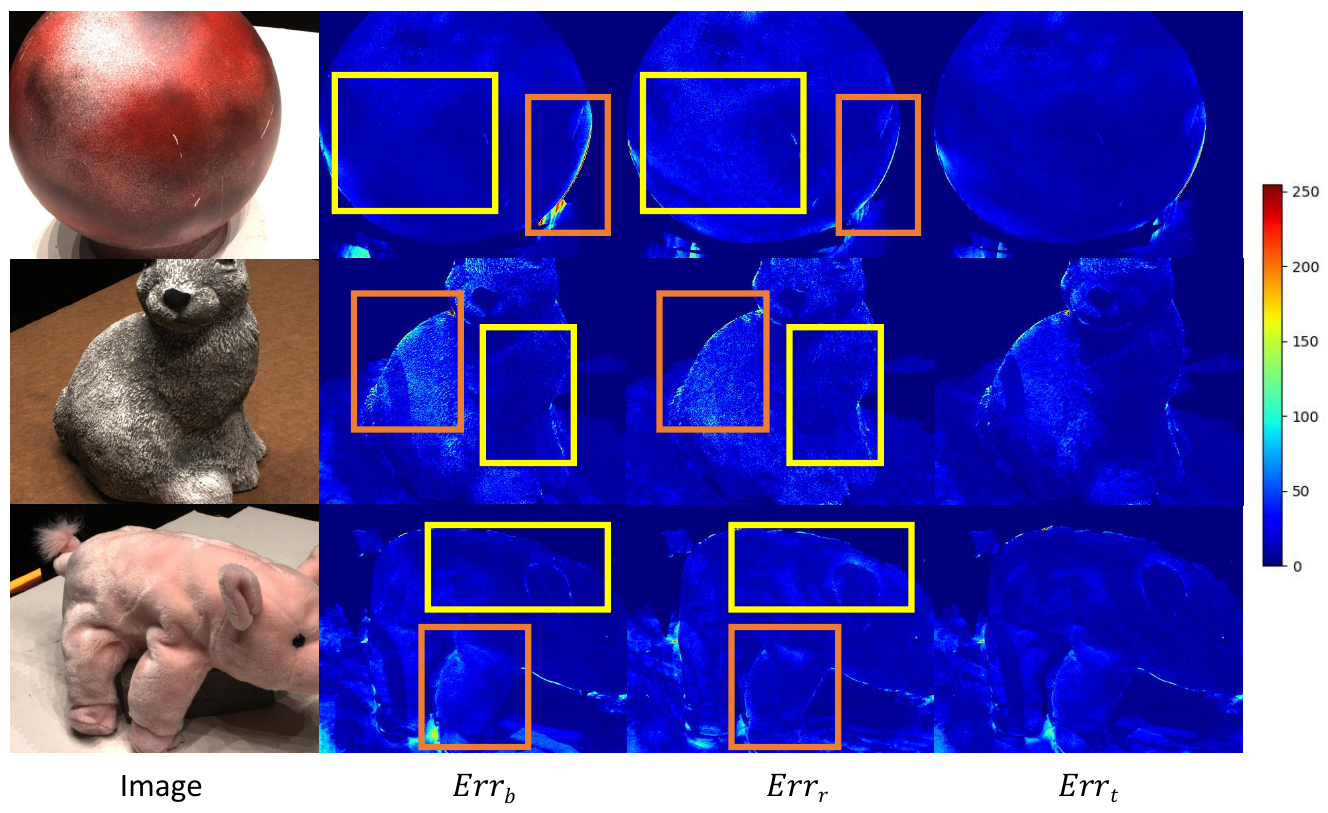}
    \caption{\textbf{Visualization of error maps.} $Err_b$ and $Err_r$ represent the error maps of the views obtained through the blending approach and the regression approach, respectively. $Err_t$ is the error map of the final fused target view. The yellow boxes indicate that the blending approach outperforms the regression approach, while the orange boxes indicate regions where the regression approach outperforms the blending approach.}
    \label{fig:err_map}
\end{figure*}

\paragraph{Error Map Visualization.}
As shown in Fig.~\ref{fig:err_map}, we present the error maps obtained by two decoding approaches, as well as the error maps of the fused views.
The blending approach tends to exhibit lower errors in most areas, while the regression approach may have lower errors in some regions with reflections and edges.
In addition, we also present quantitative results, as shown in Table~\ref{tab:intermediate_metrics}. 
The views fused through Consistency-aware Fusion (CAF) integrate the advantages of both intermediate views, achieving a further improvement in quality.

\subsection{Per-scene Breakdown}
As shown in Tables~\ref{tab:dtu_break},~\ref{tab:dtu_break_others},~\ref{tab:nerf_break}, and~\ref{tab:llff_break}, we present the per-scene breakdown results of three datasets (DTU~\cite{dtu}, NeRF Synthetic~\cite{nerf}, and Real Forward-facing~\cite{llff}). These results align with the averaged results in the main text.

\begin{table}
\centering
\renewcommand\arraystretch {1}
\resizebox{\linewidth}{!}{
\begin{tabular}{@{}l|ccccc@{}}
\toprule
Scan & \#1 & \#8 & \#21 & \#103 & \#114 \\
\midrule
Metric & \multicolumn{5}{c}{PSNR $\uparrow$} \\
\midrule
PixelNeRF~\cite{pixelnerf} & 21.64 & 23.70 & 16.04 & 16.76 & 18.40 \\
IBRNet~\cite{ibrnet} & 25.97 & 27.45 & 20.94 & 27.91 & 27.91 \\
MVSNeRF~\cite{mvsnerf} & 26.96 & 27.43 & 21.55 & 29.25 & 27.99 \\
NeuRay~\cite{neuray} & 28.59 & 27.63 & 23.05 & 29.71 & 29.23 \\
ENeRF~\cite{enerf} & 28.85 & 29.05 & 22.53 & 30.51 & 28.86 \\
GNT~\cite{gnt} & 27.25 & 28.12 & 21.67 & 28.45 & 28.01  \\
Ours & \textbf{30.72} & \textbf{30.87} & \textbf{23.96} & \textbf{31.78} & \textbf{29.84} \\
\midrule
NeRF$_{10.2h}$~\cite{nerf} & 26.62 & 28.33 & 23.24 & 30.40 & 26.47  \\
IBRNet$_{ft-1.0h}$~\cite{ibrnet} & 31.00 & \textbf{32.46} & \textbf{27.88} & \textbf{34.40} & \textbf{31.00} \\ 
MVSNeRF$_{ft-15min}$~\cite{mvsnerf} & 28.05 & 28.88 & 24.87 & 32.23 & 28.47\\
NeuRay$_{ft-1.0h}$~\cite{neuray} & 27.77 & 25.93 & 23.40 & 28.57 & 29.14 \\
ENeRF$_{ft-1.0h}$~\cite{enerf} & 30.10 & 30.50 & 22.46 & 31.42 & 29.87 \\
Ours$_{ft-15min}$ & 31.54 & 31.41 & 24.07 & 32.97 & 30.52 \\
Ours$_{ft-1.0h}$ & \textbf{31.58} & 31.61 & 24.07 & 33.09 & 30.53    \\
\midrule
Metric & \multicolumn{5}{c}{SSIM $\uparrow$} \\
\midrule
PixelNeRF~\cite{pixelnerf} & 0.827 & 0.829 & 0.691 & 0.836 & 0.763 \\
IBRNet~\cite{ibrnet} & 0.918 & 0.903 & 0.873 & 0.950 & 0.943 \\
MVSNeRF~\cite{mvsnerf} & 0.937 & 0.922 & 0.890 & 0.962 & 0.949 \\
NeuRay~\cite{neuray} & 0.872 & 0.826 & 0.830 & 0.920 & 0.901 \\
ENeRF~\cite{enerf} & 0.958 & 0.955 & 0.916 & 0.968 & 0.961 \\
GNT~\cite{gnt} & 0.922 & 0.931 & 0.881 & 0.942 & 0.960 \\
Ours & \textbf{0.971} & \textbf{0.965} & \textbf{0.943} & \textbf{0.974} & \textbf{0.965} \\
\midrule
NeRF$_{10.2h}$~\cite{nerf} & 0.902 & 0.876 & 0.874 & 0.944 & 0.913 \\
IBRNet$_{ft-1.0h}$~\cite{ibrnet} & 0.955 & 0.945 & \textbf{0.947} & \textbf{0.968} & 0.964 \\ 
MVSNeRF$_{ft-15min}$~\cite{mvsnerf} & 0.934 & 0.900 & 0.922 & 0.964 & 0.945 \\
NeuRay$_{ft-1.0h}$~\cite{neuray} & 0.872 & 0.751 & 0.845 & 0.868 & 0.900 \\
ENeRF$_{ft-1.0h}$~\cite{enerf} & 0.966 & 0.959 & 0.924 & 0.971 & 0.965 \\
Ours$_{ft-15min}$ & \textbf{0.973} & \textbf{0.967} & 0.945 & 0.976 & \textbf{0.969}  \\
Ours$_{ft-1.0h}$ & \textbf{0.973} & \textbf{0.967} & 0.945 & 0.976 & \textbf{0.969}  \\
\midrule
Metric & \multicolumn{5}{c}{LPIPS $\downarrow$} \\
\midrule
PixelNeRF~\cite{pixelnerf} & 0.373 & 0.384 & 0.407 & 0.376 & 0.372 \\
IBRNet~\cite{ibrnet} & 0.190 & 0.252 & 0.179 & 0.195 & 0.136 \\
MVSNeRF~\cite{mvsnerf} & 0.155 & 0.220 & 0.166 & 0.165 & 0.135 \\
NeuRay~\cite{neuray} & 0.157 & 0.201 & 0.156 & 0.140 & 0.128 \\
ENeRF~\cite{enerf} & 0.086 & 0.119 & 0.107 & 0.107 & 0.076 \\
GNT~\cite{gnt} & 0.143 & 0.210 & 0.171 & 0.149 & 0.139 \\
Ours & \textbf{0.061} & \textbf{0.088} & \textbf{0.068} & \textbf{0.085} & \textbf{0.065} \\
\midrule
NeRF$_{10.2h}$~\cite{nerf} & 0.265 & 0.321 & 0.246 & 0.256 & 0.225 \\
IBRNet$_{ft-1.0h}$~\cite{ibrnet} & 0.129 & 0.170 & 0.104 & 0.156 & 0.099 \\ 
MVSNeRF$_{ft-15min}$~\cite{mvsnerf} & 0.171 & 0.261 & 0.142 & 0.170 & 0.153 \\
NeuRay$_{ft-1.0h}$~\cite{neuray} & 0.155 & 0.272 & 0.142 & 0.177 & 0.125 \\
ENeRF$_{ft-1.0h}$~\cite{enerf} & 0.071 & 0.106 & 0.097 & 0.102 & 0.074 \\
Ours$_{ft-15min}$ & 0.057 & \textbf{0.082} & 0.067 & 0.080 & 0.061 \\
Ours$_{ft-1.0h}$ & \textbf{0.056} & \textbf{0.082} & \textbf{0.066} & \textbf{0.079} & \textbf{0.059} \\
\bottomrule
\end{tabular}}
\caption{\textbf{Quantitative results of five sample scenes on the DTU test set.}}
\label{tab:dtu_break}
\end{table}

\begin{table*}
\centering
\renewcommand\arraystretch {1}
\scalebox{0.80}{
\resizebox{\linewidth}{!}{
\begin{tabular}{@{}l|ccccccccccc@{}}
\toprule
Scene & \#30 & \#31 & \#34 & \#38 & \#40 & \#41 & \#45 & \#55 & \#63 & \#82 & \#110 \\
\midrule
Metric & \multicolumn{11}{c}{PSNR $\uparrow$} \\
\midrule
NeuRay~\cite{neuray} & 21.10 & 23.35 & 24.46 & 26.01 & 24.16 & 27.17 & 23.75 & 27.66 & 23.36 & 23.84 & \textbf{29.90} \\
ENeRF~\cite{enerf} & 29.20 & 25.13 & 26.77 & 28.61 & 25.67 & 29.51 & 24.83 & 30.26 & 27.22 & 26.83 & 27.97 \\
GNT~\cite{gnt} & 27.13 & 23.54 & 25.10 & 27.67 & 24.48 & 28.10 & 24.54 & 28.86 & 26.36 & 26.09 & 26.93 \\
Ours & \textbf{30.94} & \textbf{26.95} & \textbf{28.21} & \textbf{29.87} & \textbf{28.62} & \textbf{31.24} & \textbf{26.01} & \textbf{32.46} & \textbf{29.24} & \textbf{29.78} & 29.30  \\
\midrule
Metric & \multicolumn{11}{c}{SSIM $\uparrow$} \\
\midrule
NeuRay~\cite{neuray}  & 0.916 & 0.851 & 0.767 & 0.800 & 0.812 & 0.872 & 0.878 & 0.870 & 0.927 & 0.919 & 0.927 \\
ENeRF~\cite{enerf} & 0.981 & 0.937 & 0.934 & 0.946 & 0.947 & 0.960 & 0.948 & 0.973 & 0.978 & 0.971 & 0.974 \\
GNT~\cite{gnt} & 0.954 & 0.907 & 0.880 & 0.921 & 0.893 & 0.908 & 0.918 & 0.934 & 0.938 & 0.949 & 0.930 \\
Ours & \textbf{0.986} & \textbf{0.956} & \textbf{0.954} & \textbf{0.961} & \textbf{0.966} & \textbf{0.972} & \textbf{0.963} & \textbf{0.983} & \textbf{0.984} & \textbf{0.980} & \textbf{0.980} \\
\midrule
Metric & \multicolumn{11}{c}{LPIPS $\downarrow$} \\
\midrule
NeuRay~\cite{neuray} & 0.141 & 0.161 & 0.234 & 0.225 & 0.209 & 0.172 & 0.121 & 0.163 & 0.104 & 0.119 & 0.116 \\
ENeRF~\cite{enerf} & 0.052 & 0.108 & 0.117 & 0.118 & 0.120 & 0.091 & 0.077 & 0.069 & 0.048 & 0.066 & 0.069  \\
GNT~\cite{gnt} & 0.110 & 0.172 & 0.201 & 0.231 & 0.116 & 0.168 & 0.134 & 0.155 & 0.127 & 0.138 & 0.127 \\
Ours & \textbf{0.039} & \textbf{0.075} & \textbf{0.085} & \textbf{0.082} & \textbf{0.082} & \textbf{0.065} & \textbf{0.051} & \textbf{0.045} & \textbf{0.032} & \textbf{0.044} & \textbf{0.052} \\
\bottomrule
\end{tabular}}
}
\caption{\textbf{Quantitative results of other eleven scenes on the DTU test set.}}
\label{tab:dtu_break_others}
\end{table*}

\begin{table*}
\centering
\renewcommand\arraystretch {1}
\scalebox{0.80}{
\resizebox{\linewidth}{!}{
\begin{tabular}{@{}l|cccccccc@{}}
\toprule
Scene & Chair & Drums & Ficus & Hotdog & Lego & Materials & Mic & Ship\\
\midrule
Metric & \multicolumn{8}{c}{PSNR $\uparrow$} \\
\midrule
PixelNeRF~\cite{pixelnerf} & 7.18 & 8.15 & 6.61 & 6.80 & 7.74 & 7.61 & 7.71 & 7.30  \\
IBRNet~\cite{ibrnet} & 24.20 & 18.63 & 21.59 & 27.70 & 22.01 & 20.91 & 22.10 & 22.36 \\
MVSNeRF~\cite{mvsnerf} & 23.35 & 20.71 & 21.98 & 28.44 & 23.18 & 20.05 & 22.62 & 23.35 \\
NeuRay~\cite{neuray} & 27.27 & 21.09 & 24.09 & 30.50 & 24.38 & 21.90 & 26.08 & 21.30 \\
ENeRF~\cite{enerf} & 28.29 & 21.71 & 23.83 & 34.20 & \textbf{24.97} & 24.01 & 26.62 & 25.73 \\
GNT~\cite{gnt} & 27.98 & 20.27 & \textbf{26.86} & 29.34 & 23.17 & \textbf{30.75} & 23.19 & 24.86 \\
Ours & \textbf{28.87} & \textbf{22.33} & 24.55 & \textbf{34.96} & 24.90 & 26.08 & \textbf{27.98} & \textbf{26.22} \\
\midrule
NeRF~\cite{nerf} & \textbf{31.07} & \textbf{25.46} & \textbf{29.73} & 34.63 & \textbf{32.66} & \textbf{30.22} & \textbf{31.81} & \textbf{29.49}  \\
IBRNet$_{ft-1.0h}$~\cite{ibrnet} & 28.18 & 21.93 & 25.01 & 31.48 & 25.34 & 24.27 & 27.29 & 21.48 \\ 
MVSNeRF$_{ft-15min}$~\cite{mvsnerf} & 26.80 & 22.48 & 26.24 & 32.65 & 26.62 & 25.28 & 29.78 & 26.73 \\
NeuRay$_{ft-1.0h}$~\cite{neuray} & 27.37 & 21.69 & 23.45 & 32.26 & 26.87 & 23.03 & 28.12 & 24.49 \\
ENeRF$_{ft-1.0h}$~\cite{enerf} & 28.94 & 25.33 & 24.71 & 35.63 & 25.39 & 24.98 & 29.25 & 26.36  \\
Ours$_{ft-15min}$ & 30.93 & 23.29 & 25.46 & 36.28 & 26.96 & 26.91 & 31.20 & 27.51\\
Ours$_{ft-1.0h}$ & \textbf{31.07} & 23.38 & 25.62 & \textbf{36.73} & 27.24 & 27.05 & 31.49 & 27.87  \\
\midrule
Metric & \multicolumn{8}{c}{SSIM $\uparrow$} \\
\midrule
PixelNeRF~\cite{pixelnerf} & 0.624 & 0.670 & 0.669 & 0.669 & 0.671 & 0.644 & 0.729 & 0.584 \\
IBRNet~\cite{ibrnet} & 0.888 & 0.836 & 0.881 & 0.923 & 0.874 & 0.872 & 0.927 & 0.794 \\
MVSNeRF~\cite{mvsnerf} & 0.876 & 0.886 & 0.898 & 0.962 & 0.902 & 0.893 & 0.923 & 0.886 \\
NeuRay~\cite{neuray} & 0.912 & 0.856 & 0.901 & 0.953 & 0.899 & 0.881 & 0.952 & 0.779 \\
ENeRF~\cite{enerf} & 0.965 & 0.918 & 0.932 & 0.981 & 0.948 & 0.937 & 0.969 & 0.891\\
GNT~\cite{gnt} & 0.935 & 0.891 & \textbf{0.941} & 0.940 & 0.897 & \textbf{0.974} & 0.791 & 0.874 \\
Ours & \textbf{0.971} & \textbf{0.931} & 0.939 & \textbf{0.983} & \textbf{0.956} & 0.953 & \textbf{0.980} & \textbf{0.899} \\
\midrule
NeRF~\cite{nerf} & 0.971 & 0.943 & \textbf{0.969} & 0.980 & \textbf{0.975} & \textbf{0.968} & 0.981 & 0.908 \\
IBRNet$_{ft-1.0h}$~\cite{ibrnet} & 0.955 & 0.913 & 0.940 & 0.978 & 0.940 & 0.937 & 0.974 & 0.877 \\ 
MVSNeRF$_{ft-15min}$~\cite{mvsnerf} & 0.934 & 0.898 & 0.944 & 0.971 & 0.924 & 0.927 & 0.970 & 0.879 \\
NeuRay$_{ft-1.0h}$~\cite{neuray} & 0.920 & 0.869 & 0.895 & 0.949 & 0.912 & 0.880 & 0.954 & 0.788 \\
ENeRF$_{ft-1.0h}$~\cite{enerf} & 0.971 & \textbf{0.960} & 0.939 & 0.985 & 0.949 & 0.947 & 0.985 & 0.893  \\
Ours$_{ft-15min}$ & 0.978 & 0.936 & 0.946 & 0.987 & 0.959 & 0.958 & 0.987 & 0.909 \\
Ours$_{ft-1.0h}$ & \textbf{0.979} & 0.938 & 0.947 & \textbf{0.988} & 0.963 & 0.960 & \textbf{0.989} & \textbf{0.912}  \\
\midrule
Metric & \multicolumn{8}{c}{LPIPS $\downarrow$} \\
\midrule
PixelNeRF~\cite{pixelnerf} & 0.386 & 0.421 & 0.335 & 0.433 & 0.427 & 0.432 & 0.329 & 0.526 \\
IBRNet~\cite{ibrnet} & 0.144 & 0.241 & 0.159 & 0.175 & 0.202 & 0.164 & 0.103 & 0.369\\
MVSNeRF~\cite{mvsnerf} & 0.282 & 0.187 & 0.211 & 0.173 & 0.204 & 0.216 & 0.177 & 0.244 \\
NeuRay~\cite{neuray} & 0.146 & 0.211 & 0.184 & 0.113 & 0.126 & 0.165 & 0.104 & 0.256 \\
ENeRF~\cite{enerf} & 0.055 & 0.110 & 0.076 & \textbf{0.059} & 0.075 & 0.084 & 0.039 & 0.183 \\
GNT~\cite{gnt} & 0.065 & 0.116 & \textbf{0.063} & 0.095 & 0.112 & \textbf{0.025} & 0.243 & \textbf{0.115} \\
Ours & \textbf{0.035} & \textbf{0.089} & 0.064 & 0.060 & \textbf{0.064} & 0.054 & \textbf{0.021} & 0.175 \\
\midrule
NeRF~\cite{nerf} & 0.055 & 0.101 & \textbf{0.047} & 0.089 & 0.054 & 0.105 & 0.033 & 0.263 \\
IBRNet$_{ft-1.0h}$~\cite{ibrnet} & 0.079 & 0.133 & 0.082 & 0.093 & 0.105 & 0.093 & 0.040 & 0.257 \\ 
MVSNeRF$_{ft-15min}$~\cite{mvsnerf} & 0.129 & 0.197 & 0.171 & 0.094 & 0.176 & 0.167 & 0.117 & 0.294 \\
NeuRay$_{ft-1.0h}$~\cite{neuray}  & 0.074 & 0.136 & 0.105 & 0.072 & 0.091 & 0.137 & 0.072 & 0.230  \\
ENeRF$_{ft-1.0h}$~\cite{enerf} & 0.030 & \textbf{0.045} & 0.071 & 0.028 & 0.070 & 0.059 & 0.017 & 0.183\\
Ours$_{ft-15min}$ & 0.024 & 0.080 & 0.059 & 0.028 & 0.052 & 0.044 & 0.015 & 0.181 \\
Ours$_{ft-1.0h}$ & \textbf{0.023} & 0.076 & 0.058 & \textbf{0.026} & \textbf{0.050} & \textbf{0.043} & \textbf{0.012} & \textbf{0.179} \\
\bottomrule
\end{tabular}}
}
\caption{\textbf{Quantitative results on the NeRF Synthetic dataset.}}
\label{tab:nerf_break}
\end{table*}

\begin{table*}
\centering
\renewcommand\arraystretch {1}
\scalebox{0.80}{
\resizebox{\linewidth}{!}{
\begin{tabular}{@{}l|cccccccc@{}}
\toprule
Scene & Fern & Flower & Fortress & Horns & Leaves & Orchids & Room & Trex \\
\midrule
Metric & \multicolumn{8}{c}{PSNR $\uparrow$} \\
\midrule
PixelNeRF~\cite{pixelnerf} & 12.40 & 10.00 & 14.07 & 11.07 & 9.85 & 9.62 & 11.75 & 10.55 \\
IBRNet~\cite{ibrnet} & 20.83 & 22.38 & 27.67 & 22.06 & 18.75 & 15.29 & 27.26 & 20.06 \\
MVSNeRF~\cite{mvsnerf} & 21.15 & 24.74 & 26.03 & 23.57 & 17.51 & 17.85 & 26.95 & \textbf{23.20} \\
NeuRay~\cite{neuray} & 21.17 & \textbf{26.29} & 27.98 & 23.91 & \textbf{19.51} & \textbf{18.81} & 28.92 & 20.55 \\
ENeRF~\cite{enerf} & 21.92 & 24.28 & 30.43 & 24.49 & 19.01 & 17.94 & 29.75 & 21.21 \\
GNT~\cite{gnt} & 22.21 & 23.56 & 29.16 & 22.80 & 19.18 & 17.43 & 29.35 & 20.15 \\
Ours & \textbf{22.53} & 25.73 & \textbf{30.54} & \textbf{25.41} & 19.46 & 18.76 & \textbf{29.79} & 22.05 \\
\midrule
NeRF$_{ft-10.2h}$~\cite{nerf} & \textbf{23.87} & 26.84 & 31.37 & 25.96 & 21.21 & 19.81 & 33.54 & 25.19  \\
IBRNet$_{ft-1.0h}$~\cite{ibrnet} & 22.64 & 26.55 & 30.34 & 25.01 & 22.07 & 19.01 & 31.05 & 22.34 \\ 
MVSNeRF$_{ft-15min}$~\cite{mvsnerf} & 23.10 & 27.23 & 30.43 & 26.35 & 21.54 & 20.51 & 30.12 & 24.32 \\
NeuRay$_{ft-1.0h}$~\cite{neuray} & 22.57 & 25.98 & 29.17 & 25.40 & 20.74 & 20.36 & 27.06 & 23.43 \\
ENeRF$_{ft-1.0h}$~\cite{enerf} & 22.08 & 27.74 & 29.58 & 25.50 & 21.26 & 19.50 & 30.07 & 23.39 \\
Ours$_{ft-15min}$ & 23.67 & 27.89 & \textbf{31.63} & 27.47 & 22.41 & 20.63 & 33.69 & 25.53 \\
Ours$_{ft-1.0h}$ & 23.82 & \textbf{28.09} & \textbf{31.63} & \textbf{27.66} & \textbf{22.59} & \textbf{20.80} & \textbf{33.97} & \textbf{25.54}  \\
\midrule
Metric & \multicolumn{8}{c}{SSIM $\uparrow$} \\
\midrule
PixelNeRF~\cite{pixelnerf} & 0.531 & 0.433 & 0.674 & 0.516 & 0.268 & 0.317 & 0.691 & 0.458 \\
IBRNet~\cite{ibrnet} & 0.710 & 0.854 & 0.894 & 0.840 & 0.705 & 0.571 & 0.950 & 0.768 \\
MVSNeRF~\cite{mvsnerf} & 0.638 & 0.888 & 0.872 & 0.868 & 0.667 & 0.657 
& 0.951 & \textbf{0.868}  \\
NeuRay~\cite{neuray} & 0.632 & 0.823 & 0.829 & 0.779 & 0.668 & 0.590 & 0.916 & 0.718 \\
ENeRF~\cite{enerf} & 0.774 & 0.893 & \textbf{0.948} & 0.905 & 0.744 & 0.681 & 0.971 & 0.826 \\
GNT~\cite{gnt} & 0.736 & 0.791 & 0.867 & 0.820 & 0.650 & 0.538 & 0.945 & 0.744 \\
Ours & \textbf{0.798} & \textbf{0.912} & 0.947 & \textbf{0.924} & \textbf{0.773} & \textbf{0.725} & \textbf{0.975} & 0.848 \\
\midrule
NeRF$_{ft-10.2h}$~\cite{nerf} & 0.828 & 0.897 & 0.945 & 0.900 & 0.792 & 0.721 & 0.978 & 0.899 \\
IBRNet$_{ft-1.0h}$~\cite{ibrnet} & 0.774 & 0.909 & 0.937 & 0.904 & 0.843 & 0.705 & 0.972 & 0.842 \\ 
MVSNeRF$_{ft-15min}$~\cite{mvsnerf} & 0.795 & 0.912 & 0.943 & 0.917 & 0.826 & 0.732 & 0.966 & 0.895 \\
NeuRay$_{ft-1.0h}$~\cite{neuray} & 0.687 & 0.807 & 0.854 & 0.822 & 0.714 & 0.657 & 0.909 & 0.799 \\
ENeRF$_{ft-1.0h}$~\cite{enerf} & 0.770 & 0.923 & 0.940 & 0.904 & 0.827 & 0.725 & 0.965 & 0.869 \\
Ours$_{ft-15min}$ & 0.825 & 0.930 & \textbf{0.963} & 0.948 & 0.869 & 0.785 & 0.986 & \textbf{0.915} \\
Ours$_{ft-1.0h}$ & \textbf{0.829} & \textbf{0.932} & \textbf{0.963} & \textbf{0.949} & \textbf{0.873} & \textbf{0.791} & \textbf{0.987} & \textbf{0.915}  \\
\midrule
Metric & \multicolumn{8}{c}{LPIPS $\downarrow$} \\
\midrule
PixelNeRF~\cite{pixelnerf} & 0.650 & 0.708 & 0.608 & 0.705 & 0.695 & 0.721 & 0.611 & 0.667 \\
IBRNet~\cite{ibrnet} & 0.349 & 0.224 & 0.196 & 0.285 & 0.292 & 0.413 & 0.161 & 0.314 \\
MVSNeRF~\cite{mvsnerf} & 0.238 & 0.196 & 0.208 & 0.237 & 0.313 & 0.274 & 0.172 & 0.184 \\
NeuRay~\cite{neuray} & 0.257 & 0.162 & 0.163 & 0.225 & 0.253 & 0.283 & 0.136 & 0.254 \\
ENeRF~\cite{enerf} & 0.224 & 0.164 & \textbf{0.092} & 0.161 & 0.216 & 0.289 & 0.120 & 0.192  \\
GNT~\cite{gnt} & 0.223 & 0.203 & 0.157 & 0.208 & 0.255 & 0.341 & \textbf{0.103} & 0.275 \\
Ours & \textbf{0.185} & \textbf{0.126} & 0.101 & \textbf{0.130} & \textbf{0.188} & \textbf{0.243} & 0.150 & \textbf{0.176} \\
\midrule
NeRF$_{ft-10.2h}$~\cite{nerf} & 0.291 & 0.176 & 0.147 & 0.247 & 0.301 & 0.321 & 0.157 & 0.245 \\
IBRNet$_{ft-1.0h}$~\cite{ibrnet} & 0.266 & 0.146 & 0.133 & 0.190 & 0.180 & 0.286 & 0.089 & 0.222 \\ 
MVSNeRF$_{ft-15min}$~\cite{mvsnerf} & 0.253 & 0.143 & 0.134 & 0.188 & 0.222 & 0.258 & 0.149 & 0.187 \\
NeuRay$_{ft-1.0h}$~\cite{neuray} & 0.229 & 0.173 & 0.162 & 0.209 & 0.243 & 0.257 & 0.160 & 0.208 \\
ENeRF$_{ft-1.0h}$~\cite{enerf} & 0.197 & 0.121 & 0.101 & 0.155 & 0.168 & 0.247 & 0.113 & 0.169  \\
Ours$_{ft-15min}$ & 0.156 & 0.090 & 0.069 & 0.093 & 0.123 & 0.192 & 0.052 & 0.106 \\
Ours$_{ft-1.0h}$ & \textbf{0.151} & \textbf{0.087} & \textbf{0.068} & \textbf{0.089} & \textbf{0.116} & \textbf{0.182} & \textbf{0.051} & \textbf{0.103} \\
\bottomrule
\end{tabular}}
}
\caption{\textbf{Quantitative results on the Real Forward-facing dataset.}}
\label{tab:llff_break}
\end{table*}

\subsection{Limitations}
Although our approach can achieve high performance for view synthesis, it still has the following limitations.
1) Like many other baselines~\cite{mvsnerf,ibrnet}, our method is tailored specifically for static scenes and may not perform optimally when applied directly to dynamic scenes.
2) During per-scene optimization, the training speed and rendering speed of NeRF-based methods, including our method, are time-consuming. We will explore the potential of Gaussian Splatting~\cite{3Dgaussians} in generalizable NVS to address this issue in the future.

\end{document}